\documentclass[preprint,12pt]{elsarticle}

\usepackage{amsmath}
\usepackage{amssymb}
\usepackage{mathtools}
\usepackage{fontspec}
\usepackage{unicode-math}
\usepackage{newunicodechar}
\usepackage{graphicx}
\usepackage{longtable}
\usepackage{booktabs}
\usepackage{array}
\usepackage{calc}
\usepackage{float}
\usepackage{placeins}
\usepackage[normalem]{ulem}
\usepackage{hyperref}
\usepackage{url}

\providecommand{\hl}[1]{#1}

\providecommand{\real}[1]{#1}

\newunicodechar{²}{\ensuremath{^{2}}}
\newunicodechar{Δ}{\ensuremath{\Delta}}
\newunicodechar{θ}{\ensuremath{\theta}}
\newunicodechar{π}{\ensuremath{\pi}}
\newunicodechar{φ}{\ensuremath{\phi}}
\newunicodechar{ϕ}{\ensuremath{\phi}}
\newunicodechar{𝛺}{\ensuremath{\Omega}}
\newunicodechar{ᵗ}{\ensuremath{^{t}}}
\newunicodechar{ }{\quad}
\newunicodechar{–}{--}
\newunicodechar{’}{'}
\newunicodechar{•}{\textbullet{}}
\newunicodechar{′}{\ensuremath{{}^{\prime}}}
\newunicodechar{→}{\ensuremath{\rightarrow}}
\newunicodechar{⇒}{\ensuremath{\Rightarrow}}
\newunicodechar{∈}{\ensuremath{\in}}
\newunicodechar{−}{\ensuremath{-}}
\newunicodechar{∣}{\ensuremath{\mid}}
\newunicodechar{∼}{\ensuremath{\sim}}
\newunicodechar{≝}{\ensuremath{\triangleq}}
\newunicodechar{│}{\ensuremath{\mid}}

\graphicspath{{figures/}}
\hypersetup{
  colorlinks=true,
  linkcolor=blue,
  citecolor=blue,
  urlcolor=blue
}

\journal{}

\begin{document}

\begin{frontmatter}

\title{Intrinsic Motivation in Reinforcement Learning: A Research Agenda for Adaptive Self-Organisation}

\author[singularitynet]{Anatoly Belikov}
\ead{abelikov@singularitynet.io}

\affiliation[singularitynet]{organization={SingularityNET Foundation}}

\begin{abstract} 
Biological cells can be viewed as individual, interacting agents whose collective dynamics give rise to adaptive behaviour at multiple levels of organisation, from individual cells through tissues to whole multicellular organisms. In this perspective and tutorial article we discuss whether intrinsic rewards in artificial neural systems can support adaptation, functional specialisation and higher-level self-organisation without a shared external objective. We review empowerment, curiosity, learning progress, information gain, unsupervised skill discovery, mutual information estimation and the use of world models for intrinsic reward computation. Particular attention is given to failure modes showing when such objectives do not produce sustained exploration or increasingly complex behaviour. We argue that more capable systems may require complementary objectives, communication, memory, learning at multiple temporal scales and environmental constraints. Based on this perspective, we outline three experimental directions. 
These include a resource-constrained environment in which otherwise stable behavioural attractors become unsustainable, allowing us to test whether environmental constraints can mitigate characteristic failure modes of intrinsic objectives. The network of recurrent agents with per-agent intrinsic rewards, and a hierarchical world-model agent in which exploratory motor competence develops before goal-directed behaviour. These experiments are intended to test whether intrinsic learning can lead to adaptive organisation at progressively higher levels.
\end{abstract}


\end{frontmatter}


\section{Introduction}
In this article, we consider intrinsic motivation in reinforcement learning. RL methods are well developed both theoretically and at the implementation level. They are highly effective when provided with a dense reward function. Some progress has been made in defining universal intrinsic reward functions that are applicable across different environments. Here, intrinsic means that the reward is computed only from what an agent observes during its lifetime, without using hidden properties of the environment. This is especially interesting given that nature does not define a reward function for biological organisms; instead, evolution must have led to the development of some kind of intrinsic preferences.

Existing intrinsic objectives exhibit characteristic failure modes, but some of them are naturally complementary and may mitigate one another’s limitations. We also analyse environmental constraints, such as limited energy, and structural features of biological systems, such as the existence of relatively global communication channels, that could be introduced into artificial environments to support the development and self-organisation of intrinsically motivated agents.

\subsection{Article structure and contribution:}
We propose a research direction focused on adaptability in artificial neural networks, with the goal of developing learning systems that could demonstrate adaptive behaviour similar to that observed in biological networks. We ask whether some of their functional properties---adaptation, memory, communication, functional differentiation and the formation of higher-level organisation---can emerge in non-biological recurrent neural networks.

A broader question motivating the research agenda is whether such a learning system can produce progressively more complex adaptive behaviour and organisation at higher levels of organisation, similar to the formation of multicellular organisms.

The main contribution of the article is a comparison of what existing objectives optimise, an analysis of their failure modes, and a research agenda for studying how complementary per-agent objectives may produce higher-level adaptation.

We also propose a biologically inspired recurrent architecture and a specific environment suitable for evaluating intrinsic motivation methods.

We begin with motivation from biology.
The next few sections provide a tutorial-style overview and comparison of empowerment, DIAYN (skill discovery), learning progress and information gain.

The tutorial part, together with the appendix, contains derivations of the main identities and discusses interesting corner cases, including an analysis of why meaningful intrinsic rewards may fail to produce useful behaviour.

We provide an overview of some methods of world modelling and their possible use for intrinsic reward computation, and conclude with a biologically motivated research agenda: experiments intended to improve our understanding of the intrinsic motivation methods discussed.

This article is inspired by the work of Jürgen Schmidhuber, who developed the first formal computational methods for intrinsic motivation and artificial curiosity in reinforcement learning, as well as by Michael Levin's work on adaptive behaviour in biological systems across different scales.

\section{Motivation from biology}\label{motivation-from-biology}

AI as a field often draws inspiration from biology. Recent discoveries point that there is at least some intelligence on different levels of organization. Molecular networks have memory including pavlovian conditioning, associative learning etc.

Also there is no large difference between neural cells and somatic cells. Somatic cells exhibit learning, problem solving, they also use electricity for communication.

\begin{figure}
\label{low-level-learning-bio}
\includegraphics[width=5.67708in,height=2.22917in]{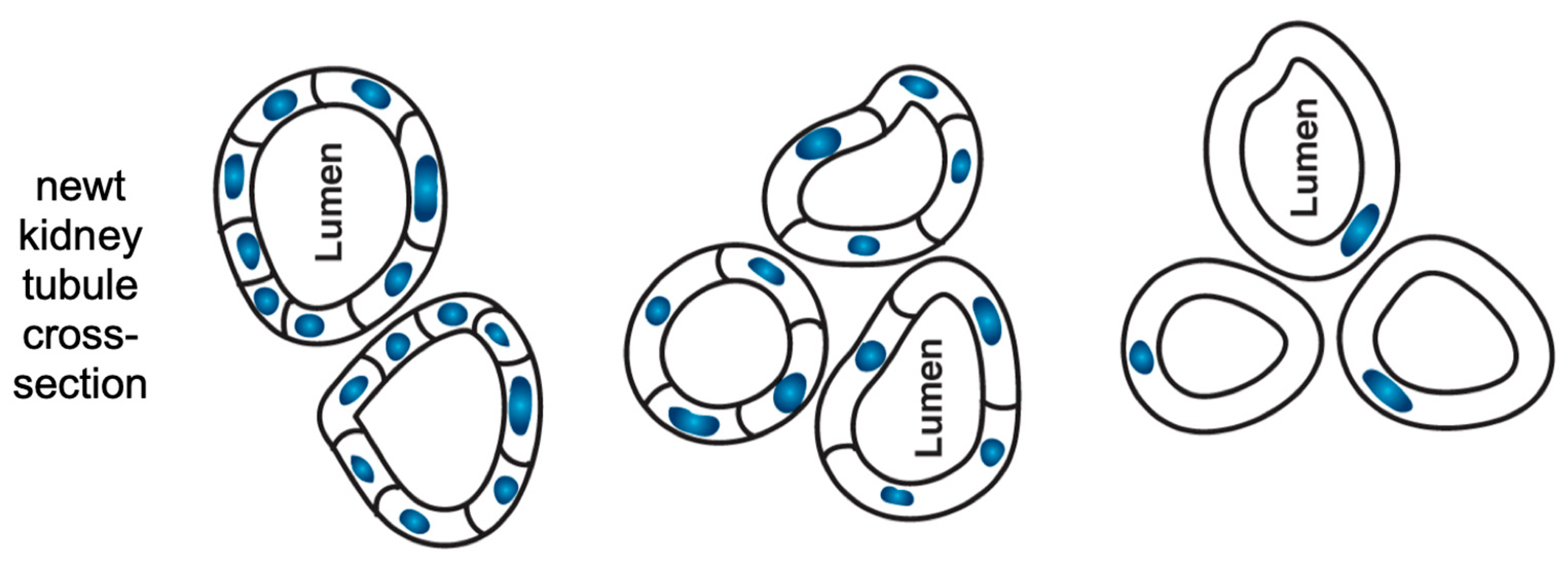}
\caption{Kidney tubules in the newt are made with a constant size, whereas cell size can vary drastically under polyploidy. The same shape achieved through different molecular mechanisms: cell to cell communication vs cytoskeleton bending. Adapted from \cite{levin2024selfimprovising}.}
\end{figure}

Some notable examples of plasticity and problem solving in image \ref{low-level-learning-bio}

\begin{figure}
\includegraphics[width=5.72917in,height=5.66667in]{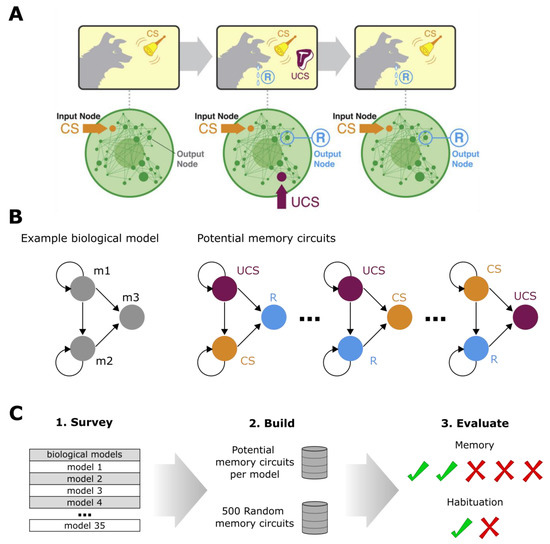}
\caption{Classical conditioning in biological network, for example drug-drug conditioning. Adapted from Figure 1 in \cite{biswas2023learning}.}
\end{figure}

\begin{figure}

\includegraphics[width=5.09375in,height=3.63542in]{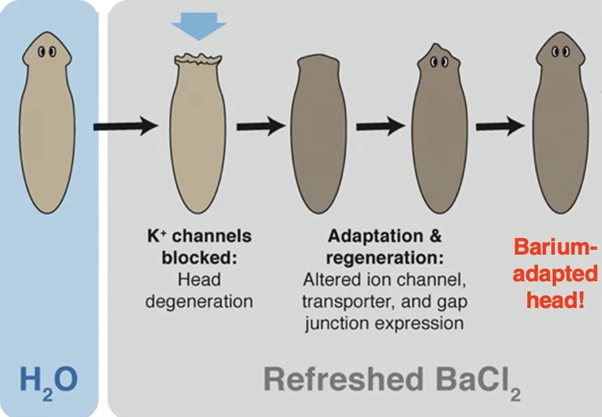}

\caption{Adaptation of neurons in planaria to barium chloride. After exposure to BaCL2 neurons in planaria's head die off. New head has resistant neurons. It's unlikely that planaria has ever been exposed to BaCL2 before. Adapted from Figure 5 in \cite{chisciure2025cognition}.}
\end{figure}

But perhaps the most striking and important case is cancer plasticity. Tumors switch between metabolic pathways (e.g., glycolysis to oxidative phosphorylation) to survive fluctuating conditions, a form of tissue-level adaptability.

For more detailed overviews, see Michael Levin's interview \emph{Michael Levin: Intelligence Beyond the Brain} and his presentation \emph{Bioelectricity: A Bridge between Physics and Cognition, by Way of Biology} \cite{levin_intelligence_beyond_brain,levin2025bioelectricity}.
\newline

\textbf{evolutionary trend hypothesis}

There appears to be a trend in growing intelligence across biosphere. Both - upper cap and average(measured by total biomass). For insects it's estimated that eusocial species make now around 50\% \cite{wilson1971insect} of insects biomass while paradoxically constituting only about 2\% of species. Modern colony size are relatively recent invention \cite{vida2025postkpg} and there is growing evidence that colony size is a primary drive of specialisation \cite{bellroberts2024evolution}. This might appear obvious in hindsight given growing specialisation of people in modern economy. This transition provides evidence that evolution may increase not only the upper bound of intelligence, but also the weighted by biomass prevalence of complex adaptive organization.

$C_{bar}(t) = \frac{\sum_i B_i(t) C_i(t)}  {\sum_i B_i(t)}$

here $C$ is intelligence measure\\
$B$ - biomass  \\
i - cells, organisms or colonies.

Possible mechanism is random specialization of some species in intelligence-based adaptation with later evolutionary arms race.

\subsection{Learning hierarchy}\label{learning-hierarchy}
We can very roughly sort different learning types from more simple to more advanced forms, that likely appeared later in evolutionary history.

\begin{enumerate}
\def\labelenumi{\arabic{enumi}.}
\item
  Non-associative learning
\end{enumerate}

Habituation - decrease of response to non-harmful repeated stimulus

Sensitization - increase of response after exposure to strong or harmful stimulus.

\begin{enumerate}
\def\labelenumi{\arabic{enumi}.}
\setcounter{enumi}{1}
\item
  Associative learning
\end{enumerate}

Classical conditioning: learning of stimulus-outcome association .

Operant Conditioning: learning of action-outcome association.

\begin{enumerate}
\def\labelenumi{\arabic{enumi}.}
\setcounter{enumi}{2}
\item
  Flexible individual ans socially mediated learning
\end{enumerate}

\begin{quote}
Metacognition. This includes capacity to model agent's own understanding e.g. being uncertain, seek more information, being able to assess own likelihood of error.

Spatial Learning \& Navigation

Insightful Problem Solving, probably based on some internal modelling as opposed to trial and error.

Teaching \& Pedagogy

Play

\end{quote}

\begin{enumerate}
\def\labelenumi{\arabic{enumi}.}
\setcounter{enumi}{3}
\item
  Cumulative Culture:
\end{enumerate}

Cultural learning over generations, learning from purely symbolic input e.g. reading instructions.

\begin{figure}
	\centering
	\includegraphics[width=6.59299in,height=4.94474in]{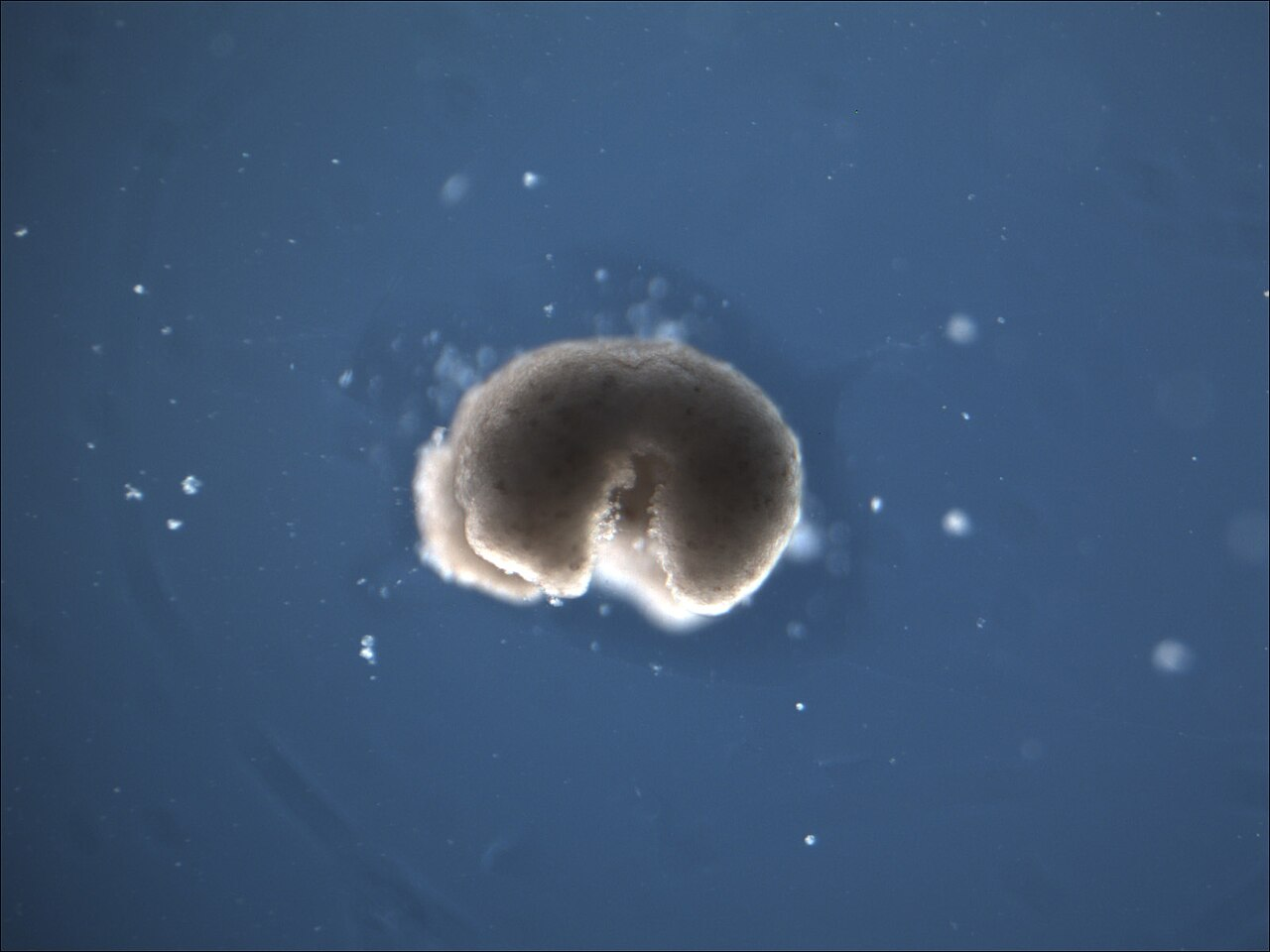}
	
	\caption{Xenobots - small, self-organizing robots made from frog embryonic cells (Xenopus laevis). This one is made of skin and cardiac cells. Image adapted from \cite{kriegman2020xenobotquadruped}, Wikimedia Commons, licensed under CC BY 4.0.}
\end{figure}

There is no universally accepted hierarchy covering all forms of learning and cognition. So this classification is rather a heuristic. But there is a good evidence that  non-associative mechanisms are evolutionarily ancient, whereas flexible planning, metacognitive control, teaching, and cumulative culture appeared much later.

Level 2 and Level 3 learning types can be observed already in insects. Bumblebee for example would play with appropriately sized balls without any external reward from only intrinsic motivation \cite{galpayagedona2022bumblebees}. Another example is Portia spiders which employ a sophisticated hunting strategy involving mimicry, detours, and ambush tactics when hunting on other spiders. Portia often attacks other spiders in their own web, so it has to be clever. Sometimes it will literally lure the prey by mimicking vibrations of an insect being stuck. It has been reported that Portia can learn to hunt a new, never seen before spider by trial and error.

Portia is a very good example since it has a very small brain. Bumblebees have approximately 950k - 1 mln neurons vs Portia around 100k.

From observations we can conclude that it has:

\textbf{Internal Representation:} The ability to take a detour where the prey is out of sight for extended periods implies the spider is not just reacting to the prey\textquotesingle s presence. It must be operating from a stored representation of the environment and the prey\textquotesingle s location within it.

\textbf{Object Permanence and Spatial Memory:} Portia\textquotesingle s behavior indicates it has a grasp of object permanence (knowing the prey still exists even when it can\textquotesingle t be seen) and strong spatial memory to navigate the planned route.

\textbf{Systematic Scanning:} The process of systematically scanning its surroundings before a hunt is interpreted as the spider building a detailed mental map of the area, which it then uses to execute its plan.

\textbf{Expectancy Violation:} When a spider takes a detour and finds wrong number of pray it spent more time inspecting the scene. This implies it had an expectation of what it should find, based on its internal model, and that expectation was violated.

Also there is growing evidence about tool use by insects. Image 5 shows the experiment: researchers gave hungry ants containers with sugar water. Researchers then altered the surface tension of the sugar water by adding surfactant. When surfactant concentrations were over 0.05\%, representing considerable drowning risk, ants were observed building the sand structures to syphon sugar water out of the container. These structures were never observed when ants foraged in containers of pure sugar water, indicating an adaptable approach to this novel tool use.

\begin{figure}[htbp]
\centering
\begin{minipage}[c]{0.48\linewidth}
\centering
\href{https://doi.org/10.1126/science.aag2360}{%
\includegraphics[width=\linewidth,height=0.18\textheight,keepaspectratio]{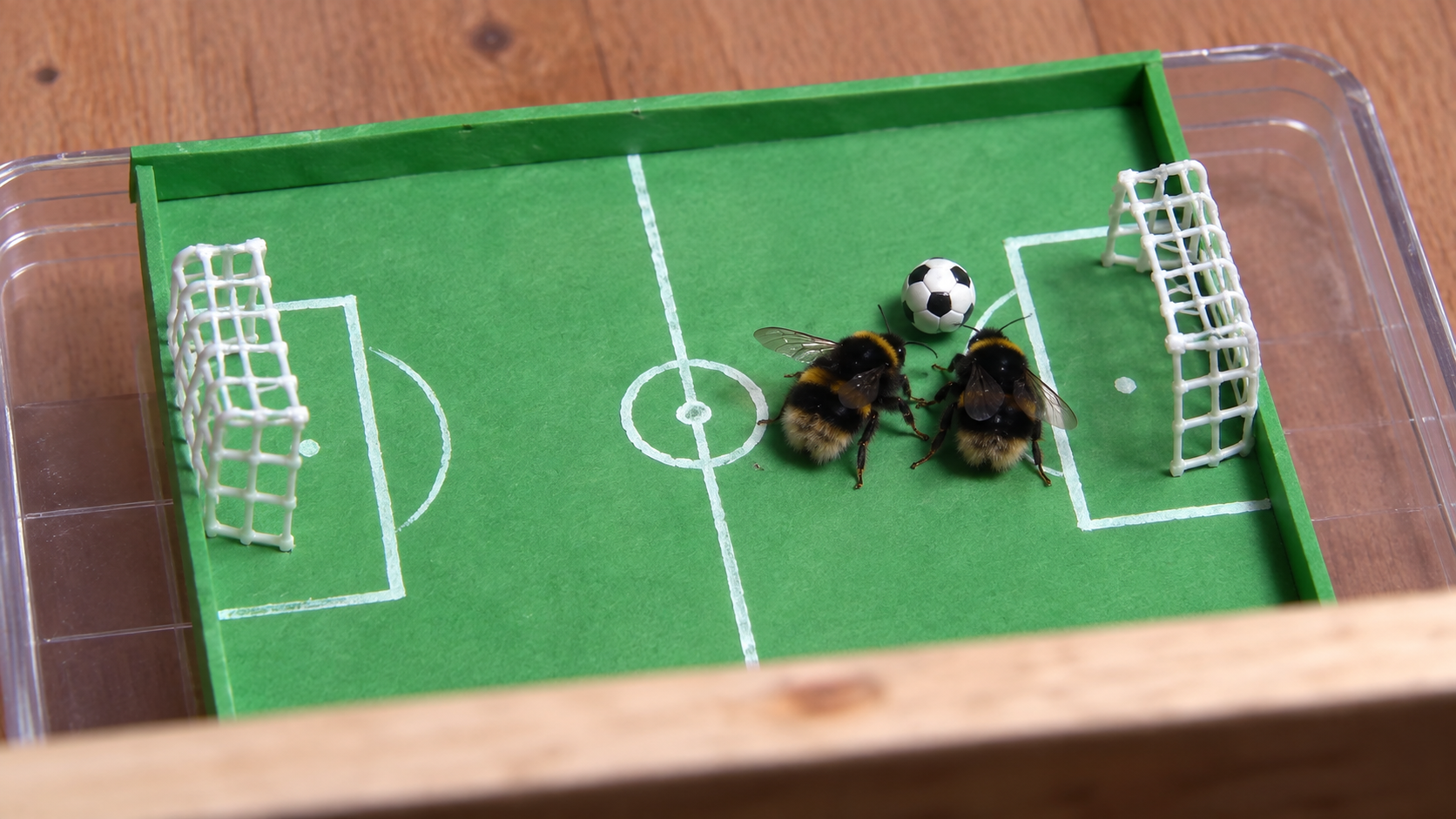}}
\end{minipage}\hfill
\begin{minipage}[c]{0.48\linewidth}
\centering
\href{https://www.sciencedirect.com/science/article/pii/S2589004222017382}{%
\includegraphics[width=\linewidth,height=0.18\textheight,keepaspectratio]{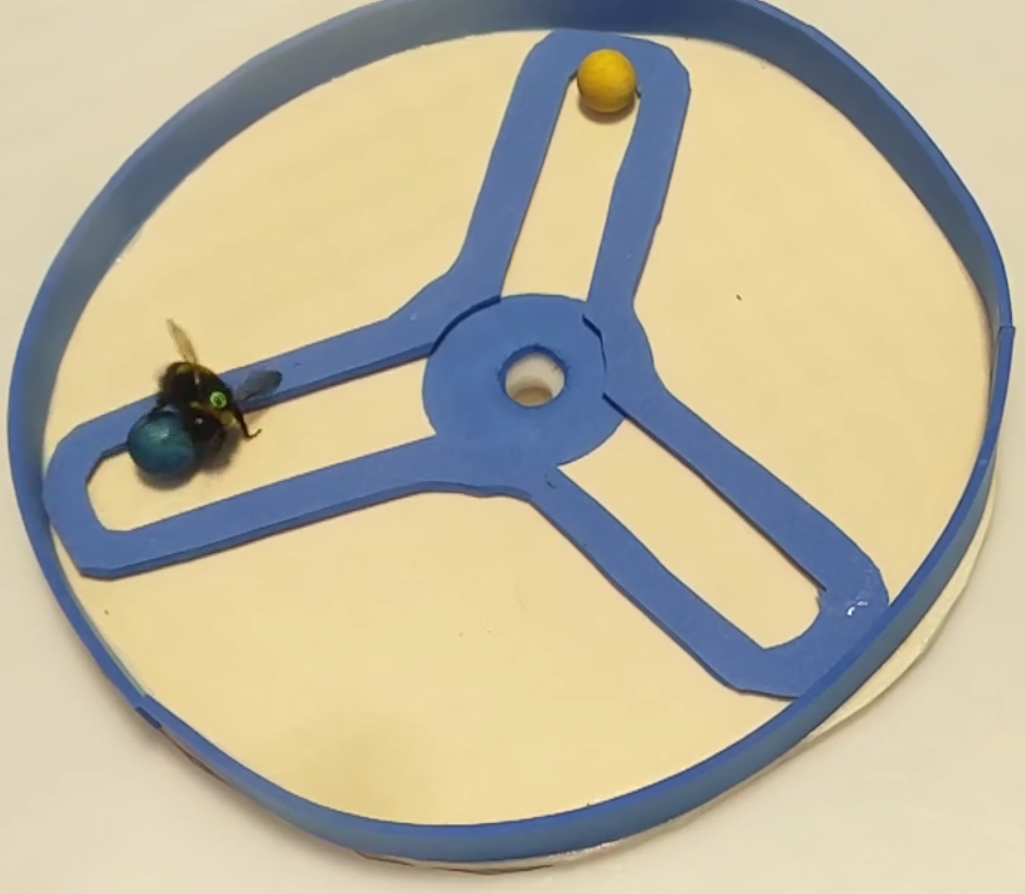}}
\end{minipage}

\medskip

\begin{minipage}[c]{0.48\linewidth}
\centering
\href{https://doi.org/10.1016/j.anbehav.2022.08.013}{%
\includegraphics[width=\linewidth,height=0.18\textheight,keepaspectratio]{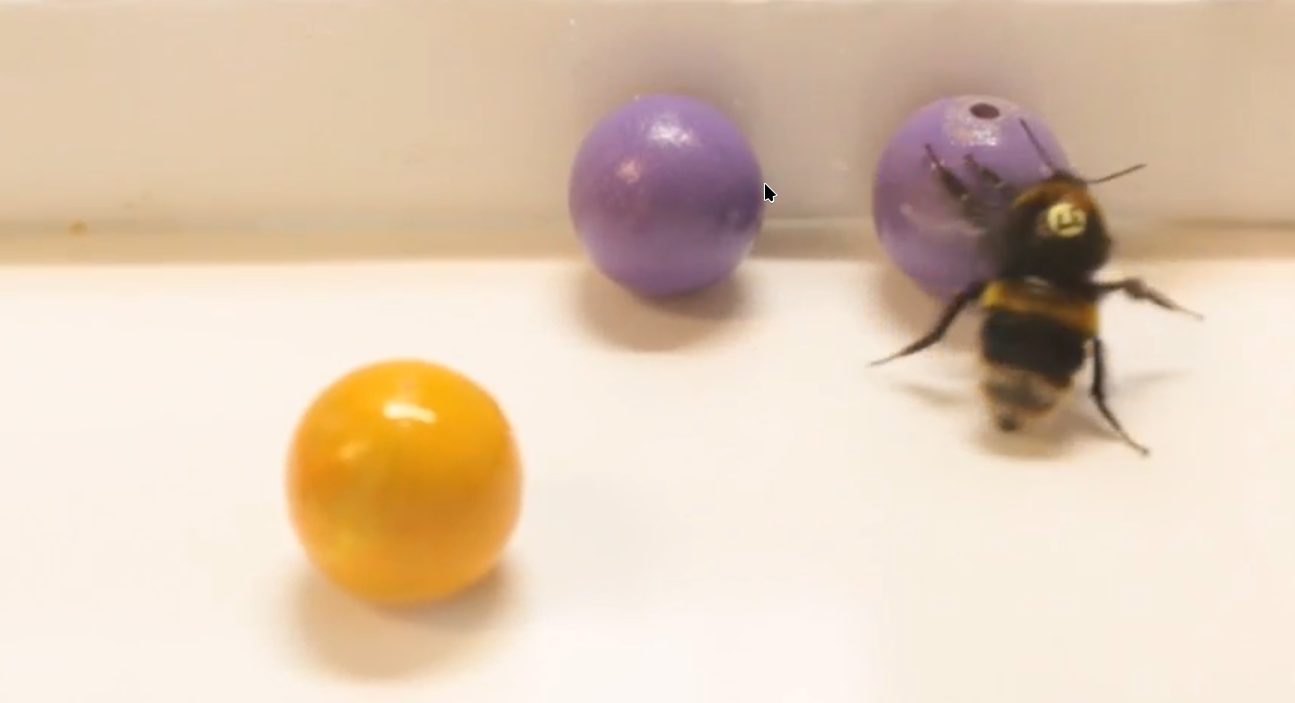}}
\end{minipage}\hfill
\begin{minipage}[c]{0.48\linewidth}
\centering
\href{https://www.inaturalist.org/observations/187462941}{%
\includegraphics[width=\linewidth,height=0.18\textheight,keepaspectratio]{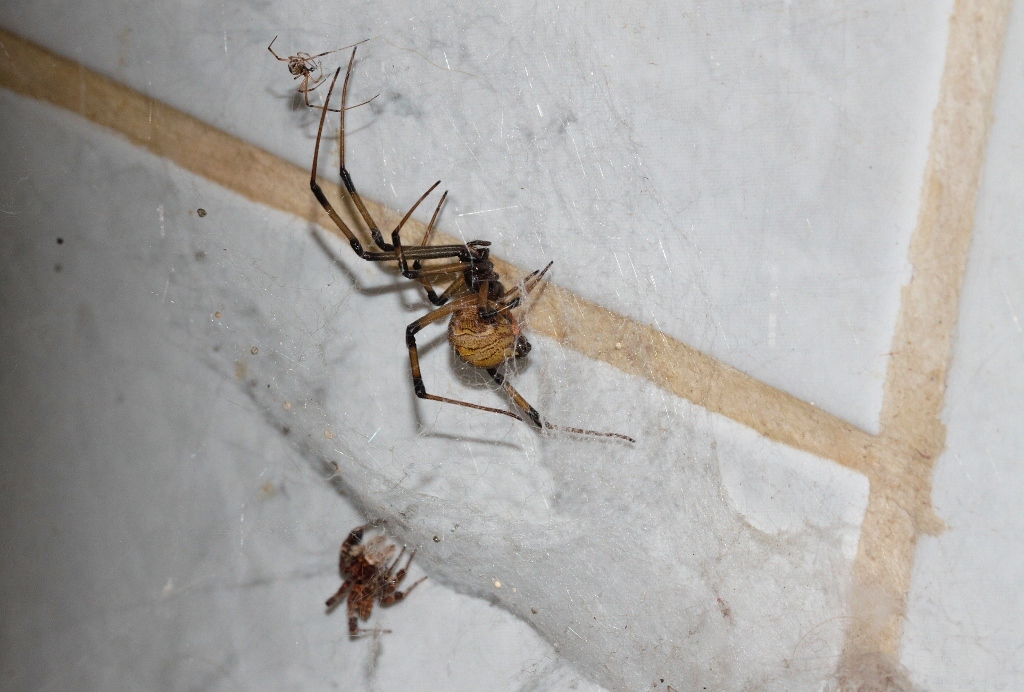}}
\end{minipage}

\medskip

\begin{minipage}[c]{0.58\linewidth}
\centering
\href{https://doi.org/10.1111/1365-2435.13671}{%
\includegraphics[width=\linewidth,height=0.20\textheight,keepaspectratio]{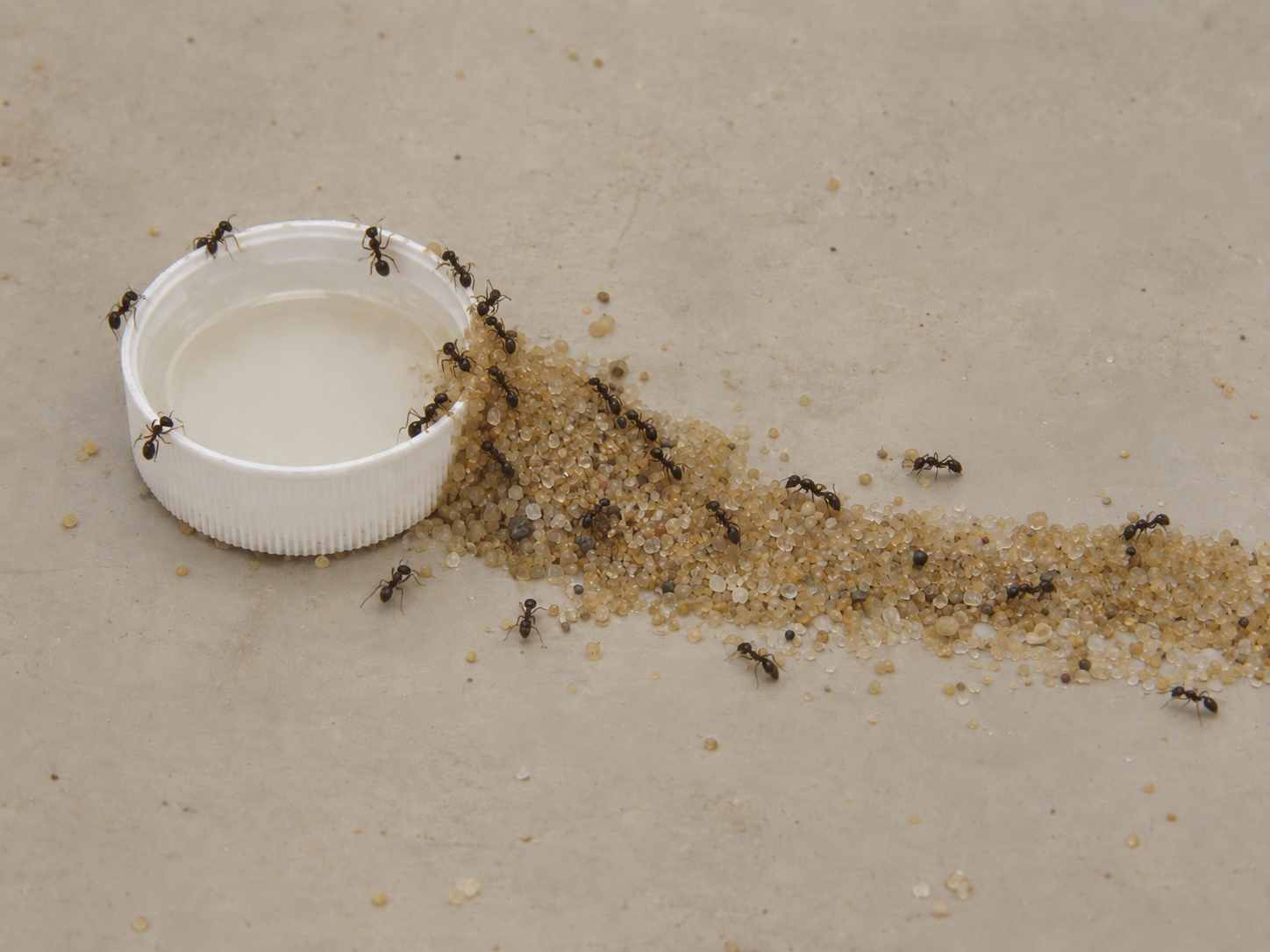}}
\end{minipage}
\caption{Examples of learned behaviour, play, complex hunting strategy and tool use in arthropods. The trained bumblebee ball-rolling image is a generated illustration of the experiment reported by Loukola et al. \cite{loukola2017cognitive}. The tool-selection image is a frame from the supplementary video by Chow et al. \cite{chow2022prior}, licensed under CC BY 4.0. The spontaneous bumblebee ball-rolling image is adapted from Galpayage Dona et al. \cite{galpayagedona2022bumblebees} under CC BY 4.0. The Portia image, ``Male Portia waiting for opportunity to predate buttonspider,'' is by i\_c\_riddell under CC BY. The ant tool-use image is a generated illustration of the experiment reported by Zhou et al. \cite{zhou2020ants}. Each image links to its source or related publication.}
\label{fig:biological-motivation-examples}
\end{figure}

\begin{enumerate}
\def\labelenumi{\arabic{enumi}.}
\item
  Biological organisms are adaptive on different levels of organization
\item
  Learn without backpropagation
\item
  All cells types can communicate with each other
\item
  Demonstrate goal-directed behaviour
\end{enumerate}

Current neural architectures are more fragile than biological ones.

Our motivation is to determine if current intrinsic motivation methods would lead to learning and behaviour types observable in insects. Whether these types of behaviours are reproducible with intrinsic rewards.

Could intrinsic motivation in RL lead to the development of level 2 and level 3 learning types:
\begin{enumerate}
\def\labelenumi{\arabic{enumi})}
\item
  Does it lead to emergent communication and cooperation?
\item
  Lead to observational learning?
\item
  Does it lead to play and episodic learning?
\end{enumerate}

\pagebreak
\section{Basic mathematical definitions}\label{basic-math-definitions}

\subsection*{Entropy}

For a discrete random variable $Y$ with probability mass function $p(y)$ we define entropy H:

\begin{equation}
	H(Y)=-\sum_y p(y)\log p(y).
\end{equation}

\begin{equation}
	H(Y|Z) =	-\sum_{y,z}p(y,z)\log p(y|z).
\end{equation}

\subsection*{Kullback--Leibler divergence}

For two probability distributions $P$ and $Q$ over the same sample space,

\begin{equation}
	D_{\mathrm{KL}}(P\|Q)
	=
	\sum_x p(x)\log\frac{p(x)}{q(x)}
	=
	\mathbb{E}_{x\sim P}
	\left[
	\log p(x)-\log q(x)
	\right].
\end{equation}

\subsection*{Mutual information}

Mutual information of two random variables Y, Z

\begin{equation}
	I(Y;Z) := D_{KL}\bigl(P(Y,Z)\,\|\,P(Y)P(Z)\bigr)
	\label{eq:mutual-information-kl}
\end{equation}

Mutual information defined via entropy

\begin{equation}
	I(Y;Z) := H(Y)-H(Y\mid Z) = H(Z)-H(Z\mid Y)
	\label{eq:mutual-information-entropy}
\end{equation}

Pointwise mutual information $ \operatorname{PMI}(x,y)
= \log \frac{p(x,y)}{p(x)p(y)},$ where x and y are events, not random variables.

It's possible for $\operatorname{PMI}$ to be negative, but it's expectation is always non-negative.

$I(X;Y)  =  {\mathbb{E}}_{(x,y) \sim p(x,y)} \lbrack \operatorname{PMI}(x,y) \rbrack$

\subsection{Reinforcement learning}

Reinforcement learning objective - maximising expected discounted reward

\begin{equation}
J(\pi) := \mathbb{E}_{p_π(τ)}\left[\sum_t \gamma^t r_t\right]
\label{eq:rl-objective}
\end{equation}
$\pi$ - policy.\\ 
$\gamma$ - discount factor in range $(0, 1]$ \\
$s_{t}$ - state at time t.\\
$p_π(τ)$ - trajectory distribution under policy. 

For a concise introduction to policy-gradient methods, see \cite{peters2010policy}.

\section{Empowerment in reinforcement learning}\label{empowerment-in-reinforcement-learning}

Mutual information between agent's action and observable states can be used as reward. In this section, we derive such a reward and discuss its relation to empowerment. Empowerment is defined as channel capacity from agent's action to subsequent state \cite{klyubin2005empowerment}:

\begin{equation}
\mathcal{E}(s) = \max_{p(a\mid s)} I(A;S'\mid S=s).
\label{eq:empowerment-definition}
\end{equation}

We can optimise policy to increase $I_{\pi}(A;S'|s)$; $I_{\pi}(A;S'|s) \leq \mathcal{E}(s)$ gives us lower-bound estimation of empowerment.

We will refer to this quantity as on-policy empowerment.

In order to gain some intuition we will consider different formulations of the same quantity:

\begin{equation}
I(A;S'\mid s)
= D_{KL}\bigl(p(a,s'\mid s)\,\|\,p(a\mid s)p(s'\mid s)\bigr).
\label{eq:empowerment-kl-form}
\end{equation}

\begin{equation}
I(A;S'\mid s) = H(S'\mid s)-H(S'\mid A,s).
\label{eq:empowerment-forward-entropy}
\end{equation}

\begin{equation}
I(A;S'\mid s) = H(A\mid s)-H(A\mid S',s).
\label{eq:empowerment-inverse-entropy}
\end{equation}

We can rewrite $I(A; S' \mid s)$  as expectation over events:

First expand $H(S' \mid s)$:

\begin{equation}
H(S'\mid s) = -\sum_{s'}p(s'\mid s)\log p(s'\mid s).
\label{eq:future-state-entropy-expansion}
\end{equation}

Since $p(s'\mid s)=\sum_a p(a,s'\mid s)$,

\begin{equation}
H(S'\mid s)
= -\sum_{s',a}p(a,s'\mid s)\log p(s'\mid s)
= -\mathbb{E}_{p(a,s'\mid s)}\log p(s'\mid s).
\label{eq:future-state-entropy-expectation}
\end{equation}

Second term $H(S' \mid A,s)$

\begin{equation}
H(S'\mid A,s)
= -\sum_{a,s'}p(a,s'\mid s)\log p(s'\mid a,s)
= -\mathbb{E}_{p(a,s'\mid s)}\log p(s'\mid a,s).
\label{eq:conditional-future-state-entropy}
\end{equation}

Substituting Eqs.~\eqref{eq:future-state-entropy-expectation} and
\eqref{eq:conditional-future-state-entropy} into
Eq.~\eqref{eq:empowerment-forward-entropy} gives

\begin{equation}
\begin{aligned}
I(A;S'\mid s)
&= -\mathbb{E}_{p(a,s'\mid s)}\log p(s'\mid s)
   +\mathbb{E}_{p(a,s'\mid s)}\log p(s'\mid a,s) \\
&= \mathbb{E}_{p(a,s'\mid s)}
   \left[\log p(s'\mid a,s)-\log p(s'\mid s)\right].
\end{aligned}
\label{eq:empowerment-log-ratio}
\end{equation}

Here, s is the current state, $S'$ is the next state and $A$ is the action taken in state s.

$A$ and $S'$ are random variables but $s$ is not.

It's important to note that removing conditioning on the current state $s$, as in $I(A;S')$ or $I(A; S)$ will lead to a different, possibly degenerate solution.

Here $I(A;S')$ is action - future state mutual information and  $I(A; S)$ - action - current state.

Maximising $H(S') - H(S'\textbar A)$ means that future state $S'$ should be predictable from action alone, and $I(A; S) = H(S) - H(A \mid S)$ means the agent should directly map current state to action. For example always choose turn left in one room and turn right in an another room. In this case $I(A;S)$ will be high, but $I(A;S'|s)$ close to zero.

What is the behaviour that is encouraged by this type of reward?

First term that is maximised $H(S' \mid s)$ is conditional entropy of future state $S'$.

Entropy is low for spiky, concentrated probability distributions and high for more even distributions.

A binary random variable with probabilities (0.5,0.5) has an entropy of one bit. A variable with 10 possible outcomes with probability $p(s_{i})\  = \ 0.1$ entropy is \textasciitilde3.32 bits.

So $H(S' \mid s)$ is high then our agent visits a large and diverse set of states. To be more precise it means for given state $s$ policy could achieve diverse set of future states. Global state visitation entropy $H(S')$ is related to conditional as $H(S')=H(S'∣S)+I(S';S)$, so $H(S')$ is at least as large as $H(S'∣S)$. What it means in practice agent could stay in an relatively small area with a lot of achievable states it could switch between.

Second term is minimised: $H(S' \mid A,s)$, this is the entropy of the future state given the current state and action. This term encourages policy to take actions that have predictable outcomes. In a deterministic environment we have $H(S' \mid A,s)$ = 0, so only the first term is optimised.

Consider this grid world, with actions up, down, left, right:

\includegraphics[width=5.27991in,height=4.40319in]{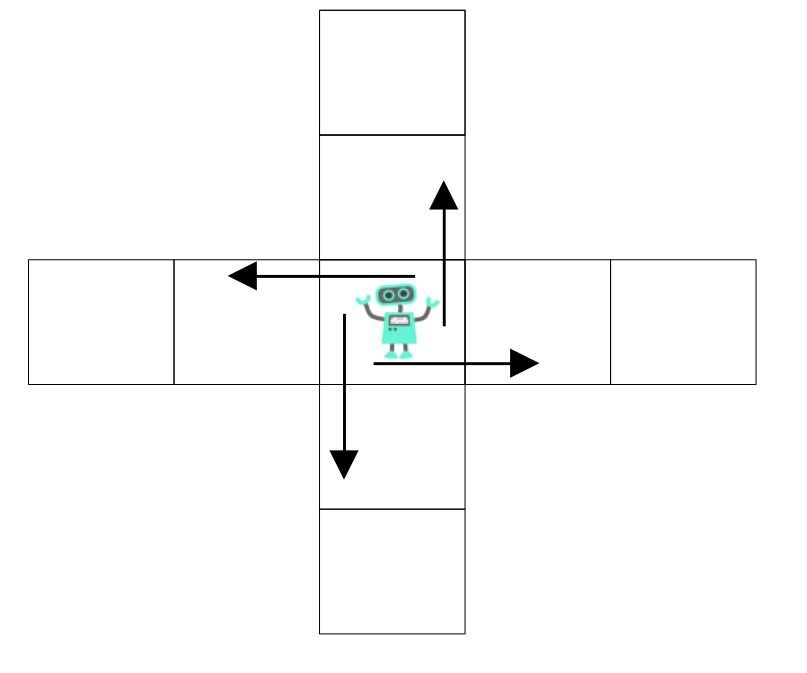}

Being in the central cell will have highest value of $H(S' \mid s)$, $- \sum_{i = 0}^{3}0.25\ log_{2}0.25\  = \ 2$ bits.

We can formalise empowerment use as intrinsic reward: $\sum_{t}^{}\gamma^{t}\ \operatorname{PMI}(s_{t + 1},\ a|\ s_{t})$, following the objective in Eq.~\eqref{eq:rl-objective}.

Since MI is an expected value of PMI it's close to original definition:

$ \mathbb{E} \left[r_t^{\mathrm{PMI}}|S_t=s
\right] = I_\pi(A;S'|S=s) \leq \mathcal{E}(s)$

Technical note: Empowerment is a property of a state, environment dynamics and the action space, not of our policy. It's already defined as maximum capacity and we can't literally optimise it. A behavioural policy may be trained both to approach this capacity and to visit states in which the capacity is high. We use the term "empowerment maximisation" for such cases.

\subsection{Inverse model trick}\label{inverse-model-trick}

We can train a model to predict the action that caused a transition from state $s_t$ to state $s_{t+1}$ and then use it as empowerment estimator. Here is how it works:

We can then estimate empowerment with equation \ref{eq:empowerment-inverse-entropy}.\\

$I(A;S'\mid s) = H(A\mid s)-H(A\mid S',s)$

By definition we have for entropies:

\begin{equation}
H(A\mid s) = -\mathbb{E}_{a \sim \pi(\cdot|s)}\log \pi(a\mid s).
\label{eq:action-entropy}
\end{equation}

\begin{equation}
H(A\mid S',s)
= -\mathbb{E}_{\substack{
      a \sim \pi(\cdot|s) \\
      s' \sim p(\cdot|a,s)
    }}\log p(a\mid S',s).
\label{eq:inverse-action-entropy}
\end{equation}

The distribution $p(a \mid s)$ is known. It is the action distribution under the current policy $\pi(a|s)$.
Distribution of actions given previous and future states $p(a \mid s',s)$ is not known, but we can approximate it with additional distribution $q(a \mid s',s)$. First we plug it in conditional entropy $H(A\mid S',s)$:

\begin{equation}
H(A\mid S',s)
= -\mathbb{E}_{\substack{
      a \sim \pi(\cdot|s) \\
      s' \sim p(\cdot|a,s)
    }}
\log\frac{p(a\mid S',s)q(a\mid S',s)}{q(a\mid S',s)}.
\label{eq:inverse-model-auxiliary-distribution}
\end{equation}

Splitting the logarithm gives

\begin{equation}
\begin{aligned}
H(A\mid S',s)
={}&-\mathbb{E}_{\substack{
      a \sim \pi(\cdot|s) \\
      s' \sim p(\cdot|a,s)
    }}\log q(a\mid S',s) \\
&-\mathbb{E}_{\substack{
      a \sim \pi(\cdot|s) \\
      s' \sim p(\cdot|a,s)
    }}
\log\frac{p(a\mid S',s)}{q(a\mid S',s)}.
\end{aligned}
\label{eq:inverse-model-log-split}
\end{equation}

The second term is, by definition, the KL divergence. Since $S'$ is itself random, this divergence must subsequently be averaged over $p(s'|s)$:

\begin{equation}
D_{KL}\bigl(p(a \mid S',s),\| q(a \mid S', s) \bigr) = \mathbb{E}_{p(s'|s)} D_{\mathrm{KL}}
\left(
p(a|s',s)\|q(a|s',s)
\right).
\end{equation}

Substitute in Eq. \ref{eq:inverse-model-log-split}:

\begin{equation}
H(A\mid S',s)
= -\mathbb{E}_{\substack{
      a \sim \pi(\cdot|s) \\
      s' \sim p(\cdot|a,s)
    }}\log q(a\mid S',s)
-D_{KL}\bigl(p(a \mid S',s),\| q(a \mid S', s) \bigr).
\label{eq:inverse-model-kl-decomposition}
\end{equation}

Substituting Eq.~\eqref{eq:inverse-model-kl-decomposition} into
Eq.~\eqref{eq:empowerment-inverse-entropy} gives

\begin{equation}
\begin{aligned}
I(A;S'\mid s)
={}&\mathbb{E}_{\substack{
      a \sim \pi(\cdot|s) \\
      s' \sim p(\cdot|a,s)
    }}\log q(a\mid S',s)
-\mathbb{E}_{a \sim \pi(\cdot|s)}\log \pi(a\mid s) \\
&+D_{KL}\bigl(p(a\mid S',s)\,\|\,q(a\mid S',s)\bigr).
\end{aligned}
\label{eq:inverse-model-bound-decomposition}
\end{equation}

p(a|s) is defined by our policy $\pi$:

\begin{equation}
\begin{aligned}
I_\pi(A;S'|s)
&= \mathbb{E}_{\substack{
      a \sim \pi(\cdot|s) \\
      s' \sim p(\cdot|a,s)
    }} \left[ \log q(a|s',s)-\log\pi(a|s) \right] \\
&+ \mathbb{E}_{p(s'|s)}
	D_{\mathrm{KL}}
	\left(
	p(a|s',s)\|q(a|s',s)
	\right).
\end{aligned}
\end{equation}

This gives us a variational lower bound on on-policy empowerment, and therefore lower bound on empowerment. This construction is similar to ELBO objective in variational autoencoder \cite{kingma2014autoencoding,doersch2016tutorial}.

\begin{equation}
I_\pi(A;S'|s)\geq
\mathbb{E}_{\substack{
  a \sim \pi(\cdot|s) \\
  s' \sim p(\cdot|a,s)
}}
\left[ \log q(a|s',s)-\log\pi(a|s) \right]
\label{eq:inverse-model-variational-bound}
\end{equation}

We can minimise divergence $D_{KL}$ term by maximising log-likelihood $\log q(a | s, s')$ on the on-policy samples $(a, s, s')$. We need only positive samples for this case, unlike MINE or InfoNCE.

For example if actions are continuous

\begin{equation}
q_{\theta}(a\mid s,s') = \mathcal{N}\bigl(a;\mu_{\theta}(s,s'),\sigma_{\theta}(s, s')\bigr)
\label{eq:inverse-model-gaussian}
\end{equation}

Then our reward function is:

\begin{equation}
r_t^{\mathrm{emp}}=\log q_{\theta}(a_t|s_t,s_{t+1})-\log\pi(a_t|s_t)
\label{eq:inverse-empowerment-reward}
\end{equation}

Our new reward function has this relation with empowerment and on-policy empowerment $I_\pi$:

\begin{equation}
\mathbb{E}[r_t^{\mathrm{emp}}|S_t=s]
\leq
I_\pi(A;S'|s)
\leq
\mathcal{E}(s)
\end{equation}

Left bound becomes tight when $q(a|s,s') = p(a|s,s')$, right bound becomes tight when policy achieves maximum capacity for given state.

We can obtain similar results for forward model that predicts future state from the equation

$I(A;S' \mid s) = H(S' \mid s) - H(S' \mid A,s)$

Though in this case we have to learn distribution of future state given current $p(S'|s)\ $which is usually a much harder problem than learning the inverse model.

In most environments there are much less options for an action that have caused transition between states $(s,s')$  than there are possible future states under $p(s'|s)$. For example in our grid world it's always only one possible action that causes transition between two cells, but there are 4 possible future states for the central cell.

The difficulty comes from simple models such as $q(s', s)  = \mathcal{N}(s'; \mu_{\theta}(s), I)$ being inherently unimodal, not suitable for modelling a multimodal distribution. It's possible to make $\mu_{\theta}(s)$ output parameter for e.g. mixture of several Gaussians, or use more advanced and difficult techniques.

\begin{figure}[htbp]
\centering
\includegraphics[
  width=\linewidth,
  height=0.75\textheight,
  keepaspectratio
]{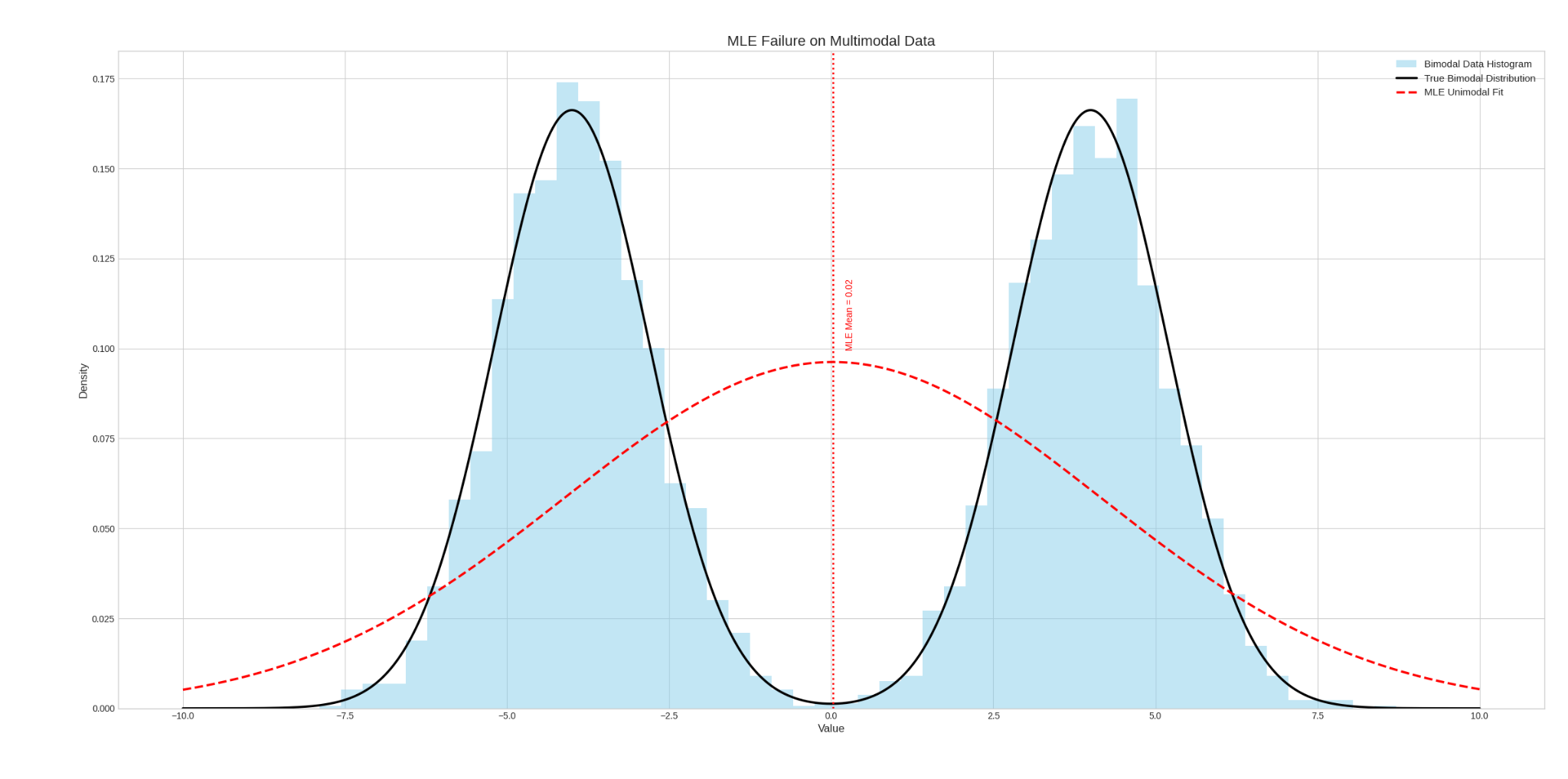}
\caption{Maximum-likelihood estimation failure with a single normal distribution.}
\label{fig:single-normal-mle-failure}
\end{figure}

A unimodal predictor will place mean between modes and give diffuse(high variance) estimation for future state S as illustrated in Fig. \ref{fig:single-normal-mle-failure}.

\subsection{Continuous(differential) entropy}\label{continuousdifferential-entropy}

Let's start with a definition of continuous entropy, it's usually noted as lower $h$ to distinguish from discrete case:

\begin{equation}
h(X)	= -\int p(x)\log p(x)\,dx
\end{equation}

Differential entropy of normal distribution:

\begin{equation}
	h(A) = \frac{1}{2}\log_{2}\bigl(2\pi e\sigma^{2}\bigr).
	\label{eq:gaussian-differential-entropy}
\end{equation}

Continuous entropy behaves quite differently from discrete, in fact for a variable $X$ with Delta-function density we have $h_{X} = - \infty$.

Mutual information on the other hand stay non-negative, but there is another issue. Consider this transition function:
$S'=A+\epsilon$

With action and noise following normal distribution:
$A\sim\mathcal{N}(0, \sigma),\qquad \epsilon\sim\mathcal{N}(0,N)$. If action reconstruction becomes possible to arbitrary precision, in such deterministic environments empowerment may become infinite.

For our inverse model we will have:

$ h(A| S', s)=\frac{1}{2}\log(2\pi e N^2)
\longrightarrow
-\infty
\qquad
\text{as }N\to 0$

\begin{equation}
I(A;S'\mid s)
= h(A\mid s)-h(A\mid S',s)
= h(A\mid s)-(-\infty)
= \infty
\label{eq:continuous-empowerment-infinite}
\end{equation}

Also for continuous actions it's possible to maximise empowerment by increasing action magnitude:

$ h(a|s) =
\frac{1}{2}\log(2\pi e\sigma^2)
\longrightarrow
\infty
\qquad
\text{as }\sigma\to \infty$

Thus, the policy will learn to maximise $h(A\mid s)$ by increasing the action range.

A natural solution is to constrain the range to $A\  \in \lbrack a_{\min},\ a_{\max}\rbrack$ or restrict covariance. To prevent entropy explosion in inverse model p(a∣s',s) we could add noise to observations or actions:

$p(a \mid s'\  + \ \epsilon,s)$ cannot perfectly recover $a$ anymore keeping entropy and mutual information finite.

For a more detailed treatment of continuous entropy and mutual information, see \cite{michalowicz2013handbook}.

\section{Multistep and single-step empowerment}\label{multistep-and-single-step-empowerment}

Consider this simple 2 step environment. Two actions are possible a0=0, and a1=1. The environment always returns 0 at t=0 and t=1, and xor of two actions at t=2:

T = 0 T = 1 T = 2, return a0 xor a1

observe 0→a0 →observe 0 → a1 → observe 1

observe 0→a0 →observe 0 → a0 → observe 0

observe 0→a1 →observe 0 → a0 → observe 1

observe 0→a1 →observe 0 → a1 → observe 0\\

The first action produces no visible one-step change:
$I(A_0;S_1∣S_0)=0$.

$O_{2}$ is independent from $O_{1}$ and $A_{1}$ and $\ I(A_{1};S_{2}|\ s_{1})  =  0$

On the other hand if agent is recurrent it(and inverse model) remembers it's actions. In such case we have $I(A_0, A_1;S_2∣S_0) = 1$ bit.

Note that final state doesn't identify each action individually:\\
$S_2=0$ implies $(A_0, A_1)∈ {00,11}$;\\
$S_2=1$ implies $(A_0, A_1) ∈{01,10}$\\

This example shows that summing local quantities,
$\sum_t I(A_t;S_{t+1}\mid S_t)$ can miss information carried jointly by multiple actions.

$I(A_{0}A_{1};S_{2} \mid s_{0})$ in our example would be two-step empowerment. We can define k-step empowerment as:

$I_{k}(s0)  \triangleq  \max_{p(a_{0:k-1}|s_0)} I(A_{0:k - 1} ; S_k \mid s_0) $

With intermediate states added it is called trajectory empowerment.

$I_{k}(s_0)  \triangleq \max_{p(a_{0:H-1}|s_0)}  I(A_{0: k - 1} ; S_{1: k} \mid s_{0})$

Knowing intermediate states makes it easier to recover actions so $I(A_{0:k−1};S_k∣S_0) ≤ I(A_{0:k−1}; S_{1:k}∣S_0)$.\\
Let's expand first equation with inverse model:

\begin{equation}
	 I(A_{0:k - 1} ; Sk \mid s_0) \geq \mathbb{E} \log q(A_{0:k−1}|S_k, s_0) − \log p_π(A_{0:k−1}∣s_0​)
\end{equation}

\subsection{Marginal and Causal objective}
We can sample all actions at s0 and then execute them one by one. In this case actions probability term is easy to compute:
$p_\pi(a_{0:H-1}\mid s_0) = \prod_{t=0}^{H-1} \pi(a_t \mid s_0, a_{0:t-1})$

Or we are sampling actions one by one, a new action at each new state. In this case to compute actions probability we need expectation over all possible trajectories between $s_0$ and $s_T$.

$p_\pi(a_{0:H-1}\mid s_0) = \int
p_\pi(a_{0:H-1}, s_{1:H-1} \mid s_0)
\,ds_{1:H-1}$

This is intractable to compute besides most simple cases. If we have a model of the environment(see sec. \ref{world-models} ) we can compute Monte-Carlo approximation by doing virtual rollouts with fixed initial state and fixed actions. Then approximation is average probability of sequence of actions over all virtual episodes. We can call it marginal MI objective since it marginalises over trajectory.

\begin{equation}
\begin{aligned}
w_j=\prod_{t=0}^{H-1}
\pi(\bar a_t\mid C^{(j)}_t) \\
\widehat p_\pi(\bar a_{0:H-1}\mid s_0)=\frac1N\sum_{j=1}^{N}w_j 
\end{aligned}
\end{equation}

Here $C^{(j)}_t$ is all relevant inputs to the policy for given episode $j$.

Another option is to plug in on-policy likelihood:

\begin{equation}
I(A_{0:k - 1} ; Sk \mid s_0) \geq \mathbb{E} \log q(A_{0:H−1};S_H| s_0) − \sum_{t=0}^{H-1}\log \pi(A_t\mid S_{0:t},A_{0:t-1})
\end{equation}

This objective can be named causal since it uses causal action entropy $\sum_t H_{\pi}(A_t|S_t, A_{i<t})$ term.

\subsection{cards and notebook environment}

Consider k-step environment:

At each time dealer draws a random card $C_t$ and stacks it on the table. The robot performs action $A_t = C_t$ such as:

\begin{itemize}
	\item wave hand left if the card is red and right if card is black
	\item write what card it sees to the notebook
\end{itemize}

After H cards are drawn it's possible to examine stacked cards and determine which actions the robot took. There are total $2^k$ possible actions sequences. So we will get up to k bits of information\\

$\begin{aligned}
	r
	&=
	\log q(\bar a_{0:k-1}\mid S_k, s_0)- \log \widehat p_\pi(\bar a_{0:k-1}\mid s_0)\\
	&\longrightarrow
	0-\log 2^{-H}\\
	&=H\log 2.
\end{aligned}$

On other hand entropy term in causal objective is always zero since each action is determined by the card. So copying environment is not rewarded. Thus marginal objective $\geq$ causal. It's important to note that causal objective is not a mutual information, it can be negative. Consider our card environment but this time cards are removed after being shown to the robot, $S_2=blank$. Policy still copies actions so $π(A_1=C| S_1=C) = 1$. Then causal entropy is $H(A_1∣S_0,S_1)=0$. Inverse model can't reconstruct $A_1$ from $S_0,S_2$, so it assigns equal probability 1/2 and $\log_2 π(A_1∣S_0,S_1) = -1$. Thus we have $I_causal = -1 - 0 = -1$.

On other hand for an agent with a notebook maximum is achieved with uniform policy $H(A_t∣S_t)=\log K$ and $J=k \log K$. Thus the policy might learn to write down random symbols to notebook. Similar failure modes exist for all information-based rewards. This issue is discussed in more details in the next section. 

We can also sample so-called latent plan $U$, or skill at the begging and optimise for this objective 

with objective
$\mathcal{E}^{U}_H(s_0) = \max_{p(u|s_0)} I(U;S_H|s_0)$

with variational lower bound

$I(U;S_H\mid s_0) \ge \mathbb{E}\left[\log q(U \mid S_H,S_0) - \log p(U \mid s_0) \right]$

This option is discussed in sec. \ref{diversity-is-all-you-need} and \ref{problems-with-empowerment-and-diayn}.

\section{Local Optima}\label{local-optima}

Many people enjoy computer games, games can even hook some humans into addictive patterns. Empowerment is suitable for modelling this effect. The reward is high where there are many visited states, but the environment is controllable. An agent might get obsessed with flipping a switch that has a lot of settings because it's easy to manipulate and offers immediate, predictable feedback. Prediction-error driven agents can get stuck observing a noise source. Similarly, empowerment-driven agent can stuck interacting with a controllable objects with many states.

This problem arises for all information-based rewards and it's likely not solvable on agent's side.. People like playing video games, but even ones with addiction won't stick there because of resource constraints. Any real environment has resource constraints and energy and resource sources. We can introduce such constraints into virtual environments.

Lets add to the state energy coordinate \hl{\emph{e}∈{[}0,1{]}}

$S  =  (s, e)$

Every action consumes $\Delta e>0$. When $e < e_{crit}$ the motor controller becomes noisy or certain actions are disabled. Formally, p(s'∣s,a) grows broader ⇒ H(S'∣a,s) increases, thus lowering empowerment. There are recharge states $s_{charge}$ (pick-ups, charging pads, food) that reset \emph{e} upward.

Therefore empowerment maximiser can obtain an intrinsic drive to:

\begin{enumerate}
	\item stay away from low-energy states
	\item periodically reach recharge regions
	\item optimise its action sequence for long-term discounted reward
\end{enumerate}

I.e. the agent behaves like a living organism: forage → play/act → return to base → repeat.

It is easy to introduce to an environment with direct biological analogy, like predator-prey, but harder for other cases. On the other hand in a multi-agent world it might not be necessary. Other agents can provide a natural brake against "camping on a toy" behaviour.

\begin{itemize}
\item  Disturbance: other agents policy injects variability into transition probability $p(s_{t + 1}|\ a_{t},\ s_{t})$ leading to increase of entropy of future state $H(S' \mid A,s)$ conditional on action.

\item resource competition: agent must keep discovering new high-control niches or defend the old one.

\item Non-stationary dynamics: camping behaviour might become impossible as time goes due to the non-stationary environment itself.
\end{itemize}

When the multi-agent cure might fail:
\begin{itemize}
	\item Predictable opponents\\
	If other agents adopt highly regular or submissive policies $H(S'∣A,s)$ can still be low → empowerment can stay high.
	
	\item Collusion / territorial partition\\
	Agents may implicitly agree on ``you keep toy \#1, I keep toy \#2''. Each finds a private controllable niche. This has analogies in biology.
\end{itemize}

Other agents can destabilise simple controllable niches, but they can also create new intrinsically rewarding attractors. Their effect therefore depends on both their behaviour, environment constraints and the intrinsic objective.

\section{Communication and collective intrinsic motivation}\label{empowerment-in-communication}

Communication extend an agent's action channel beyond its own actuators. Other agents may disrupt simple controllable niches, but communication also creates a new controllable subsystem.  Empowerment can therefore be defined not only through direct physical actions, but also through causal influence mediated by other agents. As mentioned in seq. \ref{motivation-from-biology} communication is ubiquitous in multicellular organisms so it's important to consider how it affects agents with intrinsic motivation.

In the simplest case there are two agents $A$ and $B$ that can send messages to each other from some alphabet M. They take turns: $A$ sends a message and waits for a response from $B$. Then $B$ sends its message and waits for response. High empowerment reward is achievable for both agents in such a situation. Agent $A$ selects message m randomly and sends it to $B$. If $B$ responds predictably, e.g. $f(m)=(m+1)\bmod|\mathcal{M}|$ then we have high accuracy for $P(S' \mid m,s_0)$ but also high $H(S' \mid s_0)$ (since response can't be predicted without knowing the message). Since next state now is completely predictable we get empowerment equal to $H(A \mid s)$. In this case both agents will achieve maximum reward with uniform distribution over M.

Achievable empowerment is log\textbar A\textbar{}

Thus agents may obtain intrinsic reward by exchanging arbitrary messages without producing useful collective behaviour. \\
Prediction error could reward unexpected messages.\\
Learning progress - learning of changes in other agent's policy.\\
Information gain - messages exposing other agents internal state\\
diversity objectives - role and protocol specialisation.

With a shared channel in which simultaneous messages interfere, agents may develop turn-taking schedules. But with energy constraints agents might refuse sending a response, lowering other agent's empowerment.

We can think of different levels and objectives in such systems:\\
\begin{itemize}
	\item self, or per-agent empowerment - increasing each own control
	\item transfer empowerment - increasing control of other agents
		\item assistance empowerment - increasing own influence on other agents \cite{du2020ave}
	\item joint empowerment - group of communicating agents constitute an organism, or meta-agent and can optimise their joint objective
\end{itemize}

Joint empowerment is not automatically optimised when every agent independently optimises its own empowerment.
From biology we can guess that local self-empowerment may lead to competition, domination, no-communication and possibly even to organism level cooperation. Communication predates complex multicellular organisms, but the emergence of integrated multicellular organisms required communication to be combined with adhesion, functional differentiation and mechanisms that limit conflict between cells. The long evolutionary delay, possibly billions of years, between early life and complex multicellular organisation suggests that local communication and adaptive behaviour are not by themselves sufficient to produce a stable higher-level organisation. Synchronisation of organism state across many cells using signals between adjacent cells may face scaling limits due to delays and error accumulation.

There are examples of relatively broad low-capacity channels in biology. Quorum sensing(QS) is a widespread mechanism of cell-to-cell communication and coordination using signalling molecules. It is based on release of signal molecules to the outside of cells. The phenomenon has not only been described between cells of the same species (intraspecies), but also between species (interspecies) and between bacteria and higher organisms (inter-kingdom) \cite{diggle2007evolutionary}. Hormones are later development of the same mechanism.

There's also communication based on electricity. Both cell-to-cell using ion pumps embedded in cell membranes \cite{prindle2015ion}. And using relatively broad, macroscopic electric fields. Important example is electric gradient that guides body formation in  embryogenesis. Optical signalling may represent another possibility, although its functional role in cell-to-cell communication remains less  established \cite{bodis2026ultraweak}.

We can conclude that global communication channel might enable more complex behaviour development in agents optimising self-empowerment.
Analogously to chemical communication we can introduce global message $g_t$ that aggregates individual messages:
 
$g_t = \frac{1}{N} \sum_{i=1}^{N}m_{i,t}$

with aggregated message broadcasted to all agents:

$o_{i,t+1} = \left(o^{\mathrm{local}}_{i,t+1}, g_t\right)$

\section{MINE and InfoNCE}\label{mine-and-infonce}

\subsection{Mutual information neural estimation is defined as this objective:}\label{mutual-information-neural-estimation-is-defined-as-this-objective}

Mutual Information Neural Estimation (MINE) \cite{belghazi2018mine} uses the Donsker--Varadhan representation to estimate mutual information with a neural network.

\begin{equation}
	I(X;Y)
	=
	D_{\mathrm{KL}}
	\left(
	P_{XY}
	\parallel
	P_X \otimes P_Y
	\right)
	=
	\sup_{T:\Omega\rightarrow\mathbb{R}}
	\left[
	\mathbb{E}_{(X,Y)\sim P_{XY}}
	\left[T(X,Y)\right]
	-
	\log
	\mathbb{E}_{\substack{
			X\sim P_X\\
			Y\sim P_Y
	}}
	\left[e^{T(X,Y)}\right]
	\right].
\end{equation}

Here first expectation is over joint distribution and second is over marginal distributions of X and Y.

T is a function returning real number $T(X, Y)\  \rightarrow \ R$. We approximate T by neural network $T_{\theta}(X, Y)$ and use gradient ascend to find supremum.

At training time we just need Monte-Carlo samples for both terms.\\
Loss is $\widehat I_{\mathrm{MINE}}
=
\frac{1}{N}\sum_{i=1}^N T_\theta(x_i,y_i)
-
\log
\left[
\frac{1}{N}\sum_{i=1}^N
\exp T_\theta(\widetilde x_i,\widetilde y_i)
\right]$

$\widetilde{x}$, $\ \widetilde{y}$ - samples from marginal distributions, to obtain them we can draw separate batches for x and y or just shuffle one of them e.g. y from the same batch that is used in $T_{\theta}(x,\ y)$.

It can be proofed that supremum achieved when 
\begin{equation}
T^*(x,y)	= \log \frac{p_{XY}(x,y)} {p_X(x)p_Y(y)} + C
\end{equation}

$\log \frac{P_{X,Y}\ }{Q_{X,Y}} = \log P  - log\ Q  = log\ p(X, Y)  - \log p(X)P(Y) $ is just pointwise mutual information.

For empowerment estimation we have two random variables A and S, but this is conditioned on current state s, so function T will have 3 inputs.

Plugging A and S in T gives

$T(A;S'\textbar s) = log P(A;S'\textbar s) - log (P(A\textbar s) P(S'\textbar s) + c$

$T(A;S'\textbar s) = log( \hl{p(S' \textbar A,s)p(A \textbar s)) -} log (P(A \textbar s) P(S' \textbar s) + c$

$T(A;S’|s)\  = \ log\ p(S'|A,s)\  + \ log\ p(A|s)\  - \ log\ p(A|s)\  - \ log\ p(S’|s)\  + \ c$

$T(A;S’|s)\  = \ log\ p(S'|A,s)\  - \ log\ p(S’|s)\  + \ c$

$I(A;S' \mid s) = H(S' \mid s) - H(S' \mid A,s)\  = - \ E\ \lbrack\ log\ P(S' \mid s)\rbrack\  - \ ( - \ E\ \lbrack log\ p(S’|A,\ s)\rbrack$)

$I_{A,S'\ \sim\pi}(A;S' \mid s)\  = \ E\lbrack\ log\ p(S’|A,\ s)\  - \ log\ P(S' \mid s)\ \rbrack$

Thus trained T-function directly gives a point estimate of mutual information up to an additive constant. However it wont give empowerment estimation when trained on shuffled states, actions pairs. Empowerment requires outcomes from distribution conditioned on current state $p(S'|s=s_t)$. Shuffling batch will give us $I(S';(S,\ A))$ because shuffling will approximate $S' \textasciitilde{} p(S')$ - unconditional distribution instead of $P(S' \mid s)$. Compare these equations with explicitly written expectations:

$I((S,A);S') = E_{S \sim p(s)}E_{A \sim \pi( \cdot \mid S)}E_{S' \sim p( \cdot \mid S,A)} log\frac{\ p(S' \mid S,A)}{p(S')}$

$I(S';A|s) = E_{A \sim \pi( \cdot \mid S)}E_{S' \sim p( \cdot \mid S,A)} log\frac{\ p(S' \mid S = s,A)}{p(S'|S = s)}$

With $I((S,A);S’) = I(S;S’) + I(A;S’ \mid S)$ 

Compare $I(S;S’)$ to $I(A;S' \mid s)$:

$I(S;S’)\  = \ H(S’)\  - \ H(S'|S)$

$I(A;S' \mid s) = H(S' \mid s) - H(S' \mid A,s)$

So we have conflicting term $H(S' \textbar S)$ that cancels out.

$I((S,A);S') = H(S') - H(S' \mid A,s)$

\paragraph{Policy-level failure mode}\label{policy-level-failure-mode}
Using this term as a reward is problematic. $H(S')$ is fixed for all states in a minibatch, so our reward function does not distinguish if concrete action leads to better state space exploration. With all episodes being equal in this term the only option to optimise remains inverse model. This might lead to policy visiting just a few states that are easy to predict. This is an example when an intrinsic reward may correlate positively with  exploration under the current policy, yet optimising that reward may fail to shift the policy-induced state-visitation distribution towards broader coverage.

\subsection{InfoNCE}
\label{infonce}
Another method for mutual information estimation is Information Noise-Contrastive Estimation.

\begin{equation}
	L_{\mathrm{InfoNCE}}
	=
	-\frac{1}{N}
	\sum_{i=1}^{N}
	\log
	\frac{
		\exp f_\theta(x_i,y_i)
	}{
		\sum_{j=1}^{N}\exp f_\theta(x_i,y_j)
	}.
\end{equation}

Here $(x_i,y_i)$ — positive pair, $y_j, j \neq i$, — negatives, sampled from marginal distribution.
Function $f_\theta$ is analogous to $T_\theta$ in MINE. It should return high value for samples coming from joint distribution and low values for marginal distribution. It is identical to softmax operator applied to positive and negative samples and can be viewed as a standard multi-class classification problem. This estimator is often more stable than MINE. It defines lower bound on mutual information:\\
$I(X;Y) \geq \log N-L_{\mathrm{InfoNCE}}$

\section{WORLD MODELS}\label{world-models}

World models are models that infer world state from observations and predict how this state evolves given actions. This allows to train policy in virtual episodes.

Here we describe prominent models Dreamer v3 and TWISTER and their components.

\begin{figure}[htbp]
	\centering
	\includegraphics[
	width=\linewidth,
	height=0.75\textheight,
	keepaspectratio
	]{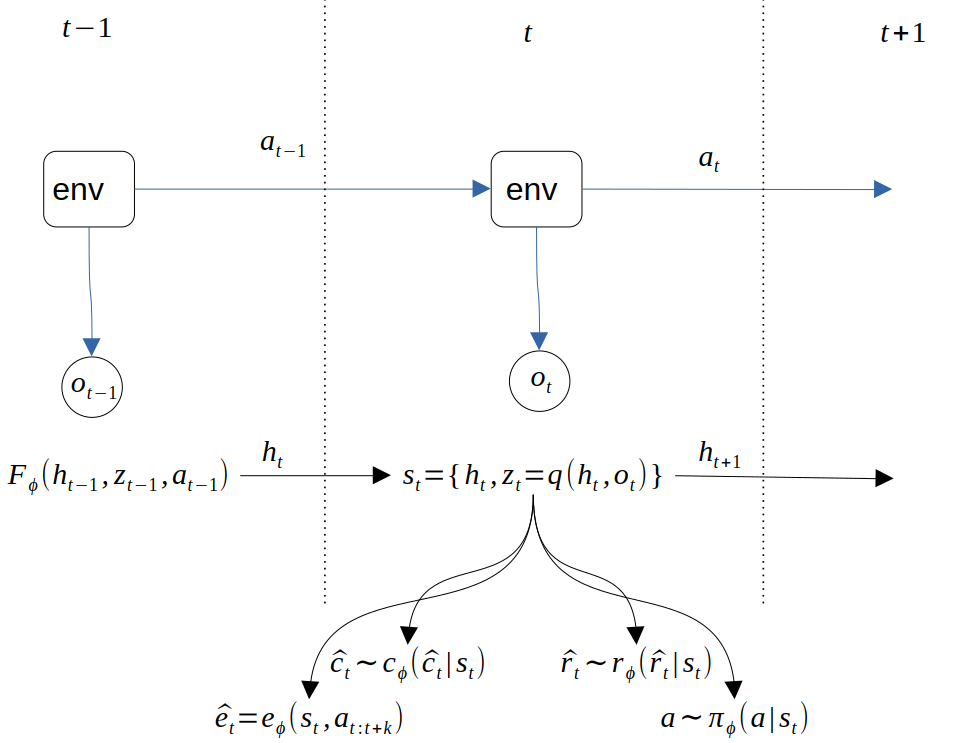}
	
	\caption{Key components of DREAMER/TWISTER}
\end{figure}

Contrastive Predictive Coding \cite{oord2018representation}

DreamerV3 \cite{hafner2023mastering}

TWISTER \cite{burchi2025learning}

Key components of DREAMER/TWISTER

$h_{t}\  = F\phi(h_{t - 1},z_{t - 1},a_{t - 1})$ - context encoder e.g. RNN or transformer.

${\widehat{z}}_{t}\  \sim \ d_{\phi}({\widehat{z}}_{t}\ |\ h_{t})$ - dynamics predictor: models distribution of world state given context. This predicts how world changes given previous state and action $a_{t - 1}$ which is also encoded in $h_{t}$

$z_{t}\  \sim \ q_{\phi}(z_{t}\ |\ o_{t},...)$ - observation encoder, models distribution of $z_{t}$ given current observation $o_{t}$ e.g. VAE encoder. In dreamer v3 its $z_{t}\  \sim \ q_{\phi}(*\ |\ o_{t},\ h_{t})$ .

$s_{t}\  = \ \{ h_{t},\ z_{t}\}$ - concatenation of context and world state.

${\widehat{o}}_{t}\  \sim \ p_{\phi}({\widehat{o}}_{t}\ |\ z_{t})$ - observation decoder, models distribution for observations e.g. VAE decoder, in dreamer it's ${\widehat{o}}_{t}\  \sim \ p_{\phi}({\widehat{o}}_{t}\ |\ s_{t})$.

${\widehat{e}}_{t} = u(z_{t})$ - model that projects world states to embedding space.

${\widehat{e}}_{t:t + k}\  = \ w_{\phi}\ (s_{t},\ a_{t:t + k})$ - predicted embeddings given state and actions

${\widehat{r}}_{t}\  \sim \ r_{\phi}({\widehat{r}}_{t}\ |\ s_{t})$ - predicted reward for current transition

${\widehat{c}}_{t}\  \sim \ c_{\phi}({\widehat{c}}_{t}\ |\ s_{t})$ - predicted episode continuation/termination

losses:

${\widehat{z}}_{t}{\sim d}_{\phi}$ is learnt with KL divergence with observation encoder $q_{\phi}(z_{t}\ |\ o_{t},...)$. 
This is very similar to recursive Bayesian estimation or Bayesian filtering, very similar to the inference in the hidden-markov model.

$q_{\phi}$ and $p_{\phi}$ are trained with whatever encoder-decoder loss is suitable e.g. ELBO loss.

$w_{\phi}$ is trained with constructive loss with projection $u(z_{t})$ of observed $z_{t}$ to predicted ${\widehat{e}}_{t:t + k}$

It is also possible to learn other useful signals such as rewards and episode termination.

\subsection{Action Conditioned (AC-CPC)}\label{action-conditioned-ac-cpc}

This is TWISTER component that is responsible for training ${\widehat{e}}_{t} = u(z_{t})$.

This is an architecture that allows compressing high-dimensional observations into useful representations.

RSSM stands for Recurrent State-Space Model. This is a part of dreamer that constitutes the world model. In TWISTER it is TSSM - transformer instead of recurrent model.

There are two key differences:

\begin{enumerate}
\def\labelenumi{\arabic{enumi})}
\item
  \textbf{Dreamer} updates distribution ${\widehat{z}}_{t}\  \sim \ d_{\phi}(*\ |\ h_{t})$ to match different encoder: $z_{t}\  \sim \ q_{\phi}(*\ |\ o_{t},\ h_{t})$.  Unlike TWISTER which uses $z_{t}\  \sim \ q_{\phi}(*\ |\ o_{t})$
\end{enumerate}

\begin{enumerate}
\def\labelenumi{\arabic{enumi})}
\setcounter{enumi}{1}
\item
  AC-CPC/TWISTER contrastive loss for embeddings additionally to reconstructive loss in Dreamer.
\end{enumerate}

Another minor difference decoder in dreamer is ${\widehat{o}}_{t}\  \sim \ p_{\phi}({\widehat{o}}_{t}\ |\ s_{t})$

Having said that it helps to concatenate many observations $o_{t}$ or use dreamer style encoder $\ q_{\phi}(z_{t}\ |\ o_{t},\ h_{t})$ for AC-CPC\hl{.}

Both methods are reported to work best with discrete distribution of world states $z_{t}$. though can be used with different distributions.

$o_0, a_0, o_1, a_1, o_2, a_2, o_3, a_3...$ - states, action sequence

$h_0=0$, posterior $z_0 = q_ϕ(h_0, o_0)$

$s_0 = [ h_0, z_0] $
$s_0 -> a_0$

$h_1=F_ϕ(z_0, h_0, a_0)$

prior $\widehat{z_1} = d_ϕ(h_1)$

posterior $z_1 = q_ϕ(h_1, O_1)$

Conceptually there information flow:  

$z_0 = q(h_0, o_0) \rightarrow h_1 = F(h_0, z_0, a_0) \rightarrow a_1$  

So $a_1$ is generated from $o_0, a_0, o_1$

$s_0$ maps to $a_0$,  

$s_1$ maps to $a_1$.

and for ac-cpc starting from h1:

$e_{10} = f(h_1)$ - conditioned by $a_0$ via $h_1$

$e_{12} = f(h_1, \widehat{ z_1}, a_1)$

$e_{13} = f(h_1,  \widehat{ z_1}, a_1, a_2)$

with positives e10 \textless-\textgreater{} z1, e12 \textless-\textgreater{} z2, e13 \textless-\textgreater{} z3\ldots{}

TWISTER applies InfoNCE objective introduced in sec. \ref{infonce} to action-conditioned predictions of future latent representations:

\begin{equation}
L_{InfoNCE}\  = \  - \frac{1}{N} \sum_{i=0}^{N} log\ \frac{e^{{s^{k}}_{ii}}}{e^{{s^{k}}_{ii}}  + \sum_{j \neq i}^{}e^{{s^{k}}_{ij}}}
\end{equation}

Here s being similarity measure between embedding and latent variable z(z is being projected to the embedding space with a multi-layer perception). N is a batch size.

Mutual information between matching embeddings e and latent states z then:

$I(e_{};\ z_{})\  > = \ log\ N\  - \ L_{InfoNCE}$

In twister this is used only for descriptor training, but what if we use this MI estimation for reward? Is it similar to empowerment? If we assume that e and z are good enough representations of it's inputs we could write:
\begin{equation}
I(e_{};\ z_{})\  \approx \ I({(h}_{t},\ a_{t:t + k}),\ h_{t + k})\  \approx \ I({(s}_{t},\ A),\ S')
\end{equation}

By chain rule for MI(see appendix)
\begin{equation}
I((s,\ A);\ S')\  = \ I(s\ ;\ S')\  + I(A;S' \mid s)
\label{eq-mi-nce}
\end{equation}

As you can see this quantity can reward behaviour when future trajectory is predictable even without knowing action.

Expand:

$I(s\ ;\ S') = \ H(S')\  - \ H(S'|s)$

$I(A;S' \mid s)\  = \ H(S' \mid s) - H(S' \mid A,s)\ $ - multistep empowerment

$H(S')\  - \ H(S'|s)$ + $H(S' \mid s) - H(S' \mid A,s)$ = $H(S') - H(S' \mid A,s)$

$I((s,\ A);\ S')$ = $H(S') - H(S' \mid A,s)$

This is difference $H(S')$ vs $H(S' \mid s)$ tells us what we are trying to distinguish/predict from s and A. AC-CPC uses for negatives all possible states S in a batch. While $H(S' \mid s)$(as in empowerment) would require us to construct negatives only from states we could arrive from the current state $s$!

Despite equation \ref{eq-mi-nce} being sound using global pool of states(or their embeddings) has issues. One issue mentioned in sec. \ref{policy-level-failure-mode} is that $H(S')$ will be the same for all episodes in the minibatch. 
Another issue is first term $I(S;S`)$ doesn't involve actions. Suppose the positive pair from a RTS-game environment is:\\
current frame: your base, five workers, daytime; \\
future frame: nearly the same scene after a few seconds.\\

A global negative might include completely different map locations, enemy type, time elapsed from start, different army and resource count. It's easy for discriminator to distinguish even without using actions. At this point $L_{nce}$ is close to zero and learning stops.
Empowerment formulation on other hand requires "hard negatives". That is states produced from the same initial condition, but under alternative actions.
The good thing is that world models allow us to generate realistic virtual episodes!

\section{Diversity is all you need}\label{diversity-is-all-you-need}

The goal of Diversity Is All You Need (DIAYN) \cite{eysenbach2018diayn} is to make an agent learn different skills in an unsupervised manner.

In this method we pass to the policy additional parameter z that is sampled from a random distribution(categorical or normal). Z ∼ p(z). Policy conditioned on z is called skill. Training objective has tree parts:

1) Mutual information between states and skills I(S;Z) is maximised

2) MI between actions and skills given the state is minimised I(A;Z \textbar{} S)

3) Entropy of actions given state is maximised H{[}A \textbar{} S{]}

\begin{equation}
\begin{aligned}
\max F(\theta) 
&\triangleq I(S;Z) + H[A|S] - I(A;Z|S) \\
&= (H[ Z] - H[ Z|S]) + H[ A|S] - (H[ A|S] - H[ A|S,Z])\\
&= H[Z] - H[Z|S] + H[A|S,Z]
\end{aligned}
\end{equation}

$H[ Z ]$ is entropy of skill distribution - constant.

Conditional entropy of skill distribution can be rewritten using chain rule for conditional entropy as:

$H[Z|S]  = H[S, Z]  -  H[S] = H[S|Z] - H[S] + H[Z]$

This will help us to understand when exactly the reward will be high and when low.

Substituting it into original equation:\\
$F(\theta) = H[Z] - (H[S, Z]  -  H[S]) + H[A|S,Z]$

$= H[Z] - H[S,Z] + H[S] + H[A|S,Z]$

Expand joint entropy of skills and states:

$H[S, Z] = H[S|Z] + H(Z) = H(Z|S) + H(S)$

$F(\theta) = H[Z] - ( H(S|Z) + H(Z)) + H(S) + H[A|S,Z]$

$= H[Z] - H(S| Z) - H(Z) + H(S) + H[A|S,Z]$

$= H(S) + H[A|S,Z] - H[S|Z]$

This form helps to understand what behaviour is encouraged by this reward.

\hl{First term \textbf{H(S)} maximises entropy of states = encourages policy to visit many states.}

$H\lbrack A\ |\ S,\ Z\rbrack$ - conditional entropy of action given state and skill - encourages policy to apply different actions given skill and state. Equivalently this means it should be hard to guess skill from action alone.

Also maximising this term makes it so knowing state + action doesn't give more information about skill than knowing just state alone.

H(S\textbar Z) is minimised. If it is low we have a small number of states which are visited by given skill.\\
\strut \\
Using definition I(Z;A∣S)=H(A∣S)−H(A∣S,Z)

\emph{H}(\emph{A}∣\emph{S},\emph{Z}) = \emph{H}(\emph{A}∣\emph{S}) - \emph{I}(\emph{Z};\emph{A}∣\emph{S})

So, $F(\theta) = \ H(S)\  + \ H(A \mid S)\  - \ H(S|Z)\  - \ I(Z;A \mid S)\ $ =

$\ H(Z) + \ H(A \mid S) - \ H(Z\ |\ S)\  - \ I(Z;A \mid S)$

Max H(S) = maximise the number of visited states.

Max H(A∣S) = maximise the number of actions taken in any particular state.

Min H(S\textbar Z) = minimise the number of states visited by a particular skill.

Min I(Z;A∣S) = all actions are more or less the same for all skills.

Min H(S\textbar Z) = skill can be used to predict S

\hl{\hfill\break
This objective can be implemented as:}

$H(Z\ |\ S)$ = $- \sum_{}^{}p(s,\ z)\ log\frac{p(s,\ z)}{p(s)} = \ E\lbrack - log\ p(z|s)\rbrack$

$H(Z)$ = $- \sum_{}^{}p($z) log p(z) = E{[}-log p(z){]}

$\ H(A \mid S,Z)$ = $- \sum_{}^{}p(s,z)\sum_{}^{}p(a|s,\ z)\ log\ p(a|\ s,z)\  = \ E_{s,z}\lbrack - \sum_{}^{}p(a|s,\ z)log\ p(a|\ s,z)\ \rbrack\ $ = $\ {- E}_{s,z,a}\lbrack log\ p(a|\ s,z)\ \rbrack$

$F(\theta)\  = \ E\lbrack - log\ p(z)\rbrack\  - \ E\lbrack - log\ p(z|s)\rbrack\  + \ E\lbrack - log\ p(a|\ s,z)\ \rbrack = \ E\lbrack\ \ log\ p(z|s)\  - log\ p(z)\ \  - log\ p(a|\ s,z)\ \ \ \ \ \ \ \ \ \ \rbrack$

Probability of skill given state p(z\textbar s) is approximated by a neural network predictor $\phi_{\theta}(s) - > z\ $

log $p(a|\ s,z)$ is just policy entropy log πθ(a\textbar{} s,z). It can be included in the reward or treated separately, like entropy reguliser as in soft actor-critic.

r(s, a\textbar{} z) = log φ(z \textbar{} s) − log p(z)

In principle it's possible to use forward density p(s\textbar z), but practically choice p(z\textbar s) is much easier to implement, since we don't know the ground truth distribution of S given skill.

Estimating φ(z∣s) is just a classification/regression problem; the target distribution over z is simple (uniform or Gaussian).

Estimating p(s∣z) is often a high-dimensional density problem (very hard for image observations).

Side by Side comparison with Empowerment

{\def\LTcaptype{none} 
\begin{longtable}[]{@{}
  >{\centering\arraybackslash}p{(\linewidth - 4\tabcolsep) * \real{0.2674}}
  >{\raggedright\arraybackslash}p{(\linewidth - 4\tabcolsep) * \real{0.3380}}
  >{\raggedright\arraybackslash}p{(\linewidth - 4\tabcolsep) * \real{0.3947}}@{}}
\toprule\noalign{}
\endhead
\bottomrule\noalign{}
\endlastfoot
& \textbf{Empowerment} & \textbf{DIAYN} \\
pushed up & \emph{p}(\emph{s`}∣\emph{a},\emph{s}) & \emph{p}(\emph{z}∣\emph{s}) \\
variance pushed up & $p(s'∣s)$ & \emph{p}(\emph{a}∣\emph{s},\emph{z}) \\
Encouraged property & Predictable consequences per action & States discriminate skills;

actions stay \emph{diverse} \\
Discouraged property & Small marginal next-state variability & Peaky action distribution inside a skill \\
\end{longtable}
}

In equation 1 variable Z is sampled ones per episode, S is picked uniformly from the whole trajectory, so the term $\ H(Z\ |\ S$) is computed with respect to the entire trajectory. But there are situations when examining multiple states might be desirable. We might be interested in achieving the same state by different means, for example robot can place a spoon in a mug either by taking a spoon and placing it or by trying to scoop the spoon. In this case intermediate states are clearly distinguish skill vector, but the end state is the same. By default DIAYN will try to avoid visiting the same state from two skills. We could counter it by either by adding external reward, curiosity, or adding structure to the skill vector as discussed below:

\textbf{Structural DIAYN}\label{structural-diayn}

For example let vector z be concatenation of vectors $z_{i}$.

If we choose to use just two vectors we will have:

\begin{itemize}
\item
  two latents z(1) and z(2) are concatenated $z = \lbrack z^{(1)},\ z^{(2)}\rbrack$;
\end{itemize}

\begin{itemize}
\item
  the policy $\pi(a \mid s,z^{(1)},\ z^{(2)})$ can use both parts all the time;
\item
  the discriminator at an early time slice tries to predict $z^{(1)}$ only, ignoring $z^{(2)}$;
\item
  the discriminator at the final state (or any late slice) predicts $z^{(2)}$ only.
\end{itemize}

Any two skills that differ only in $z^{(1)}$ therefore must converge to (almost) the same final state, reached via recognisably distinct trajectories.

We can also try to reconstruct all $z^{(i\  < \ t)}$. In this case policy will be encouraged to treat early $z^{i}\ $ as a high-level coarse plan or ``style'' and later $z^{i}$ further nuancing the behaviour. Other options such as using a sliding window are possible.

\section{Curiosity and Learning progress}\label{curiosity-lp}

Curiosity reward can be a combination of a few things: surprise, for example in the form of prediction error, novelty and learning progress. The theoretical foundations of curiosity and learning progress were developed by Jürgen Schmidhuber; see \cite{schmidhuber2010creativityweb}, \cite{schmidhuber2010formal}.

\subsection{Prediction error and Novelty }

Prediction error can be estimated from forward model:$\ f(s,\ a)\  - > \ {\widehat{s}}_{t + 1}$

Reward then is $||s_{t + 1}\  - f(s,\ a)||_{}$

Novelty is inversely related to state visitation count: it is high for new states and low for frequently visited states. It encourages large entropy H(S) of the state visitation distribution.

\paragraph{Implementation note}
In practice, novelty is often computed from embeddings of observed states. As discussed in the \nameref{policy-level-failure-mode}
paragraph and Section~\ref{world-models}, some novelty-based rewards may fail to distinguish episodes that produce broader exploration.

A second failure mode can occur when the score lacks a persistent scale across policy updates. If novelty is normalised using statistics of the current batch, small differences between increasingly similar episodes are rescaled. The normalised reward can therefore continue to rank episodes within each batch even as the absolute diversity and coverage decreases.

\subsection{Learning progress}
Learning progress is a bit more complex. We have forward model $f_{\theta}(s_{t + 1}|s,\ a)$ and distribution over its weights $p(\theta|O_{t})$ given the history of transitions; $\ O_{t}\  = \ \{(s_{\tau}\ ,\ a_{\tau}\ ,\ s'_{\tau})\}$ for $\tau < t$.

Learning progress is modeled as a change of distribution over parameters of forward model $f_{\theta}(s_{t + 1}|s,\ a)$ given a new observation.

$LP_{t} ≝ \ KL\ \lbrack\ p(\theta\ |\ O_{t}\  \cup \ O_{t + 1})\ \|\ p(\theta\ |\ O_{t})\ \rbrack$

This can be approximated simply with improvement in prediction error.

That is $r\  = ||\ f_{\theta_{k}}(s,\ a)\  - \ s_{t + 1}||_{}\ \  - ||_{}f_{\theta_{k + 1}}(s,\ a)\  - \ s_{t + 1}||_{}$

Novelty and prediction error, unlike learning progress are prone to noise-staring behaviour. That is, the reward is high when the agent observes pure noise. But there are easy workarounds. We can train a model that predicts action given previous and current states. $g_{\theta}(s_{t + 1},\ s_{t})\  - > \ a$. Higher layers of this model can be used as feature extractors, that keep information only about controllable aspects of the environment. These features can be used then instead of raw states in novelty or surprise rewards.

Let's compare curiosity with empowerment.

Empowerment I(A;S'∣s)=H(S'∣s)−H(S'∣s,a)

Prediction error H(S'∣s,a)

State novelty H(S)

Given identity $H(S') = I(S';\ S) + H(S' \mid S)$ we have H(S'∣s) is less or equal to $H(S')$.

That is possible to have large $H(S')$ and but small $H(S' \mid S)$. Reverse is not true:

High $H(S' \mid S)$ implies that $H(S')$ at least that large.

So maximising $H(S')$ is not identical to maximizing $H(S' \mid S)$. $H(S' \mid S)$ encourages rewards states from which many different states can occur immediately(or in k-steps with k-steps empowerment). But $H(S')$ rewards visiting the whole state space.

\section{Information gain}\label{information-gain}

Assume our world model estimates world state with z and tracks history in h. Then we can define (point) information gain about world state as \\
$KL(q(z \mid o, h) \parallel p(z \mid h)) = I_{pmi}(o; z \mid h)$

That is how much have we learned about world given a new observation o.

Note: term "information gain" is applicable to different things, including model parameters. In this case we have change in distribution over model parameters which is directly related to learning progress.

This quantity could be used as addition to prediction error. We have this relation:

$ H(O \mid h) = E_{z \sim p(z \mid h)}[H(O \mid z,h)] + I(O;Z \mid h)$

The first quantity is unreducible or so called \textbf{aleatoric} uncertainty. This is uncertainty that remains even if we have good state estimation. The second term is the uncertainty caused by not knowing z. If agent encounters a source of noise prediction entropy(and error) will be high, but information gain small hinting at large aleatoric uncertanty. On other case consider agent exploring unknown part of the map, in this case prediction error will be high, but information gain also large. We could use this to reward useful exploration much more than just watching random events.

Similar to other information-based rewards IG is prone to "camping on a toy" issue. For example with fair dice we have $I(O;Z|H) = H(Z|H) - H(Z|O,H) = ln(6) - 0 = ln(6)$ nats for each throw.

\section{SFA}\label{sfa}

Invented by Laurenz Wiskott and Terrence Sejnowski \cite{wiskott2002slow}, Slow Feature Analysis is an unsupervised learning rule that extracts features whose values change as slowly as possible over time, although they are computed from an input stream that may itself vary quickly.

Suppose that an encoder neural network transforms each observation $o_t$ into a feature vector $y_t$:

$y_{t} = g(o_{t})$

Slowness is achieved with loss
\begin{equation}
	L_{\mathrm{slowness}}
	=
	\frac{1}{T-1}
	\sum_{t=2}^{T}
	\|y_t-y_{t-1}\|_2^2.
\end{equation}

In order to avoid degenerate solutions such as encoding each observation as zeros we require decorrelation, zero mean and unit variance for y.

Variance loss is defined as:

$C_{Y} = \frac{1}{N} Y^{T}Y$ where Y is a concatenation of zero-centred vectors y.

\begin{equation}
	L_{\mathrm{variance}}
	=
	\frac{1}{d}
	\sum_{j=1}^{d}
	\left(C_Y[j,j]-1\right)^2.
\end{equation}

The correlation loss is the normalised squared Frobenius norm of the off-diagonal part of $C_Y$:

\begin{equation}
	L_{\mathrm{correlation}}
	=
	\frac{1}{d(d-1)}
	\sum_{i\neq j} C_Y[i,j]^2.
\end{equation}

Then objective is

$L_{sfa}\  = \ L_{slowness}\  + \ L_{variance\ }\  + \ L_{correlation}$

This is potentially useful augmentation for action conditioned state embeddings. AC-CPC in TWISTER must store relatively fast details needed for next observation prediction. We could extract slow-changing features from action-conditioned embeddings. Action conditioning is important: it may help distinguish controllable features from features that do not depend on the agents actions. Applying SFA on top of AC-CPC might therefore extract slow, controllable features suitable for computing of intrinsic reward over longer time scales.

\section{Predictive information bonus and MDL}\label{predictive-information-bonus-and-mdl}

Imagine an agent that can see part of an image, can move on it and can change pixels. Empowerment alone won't produce interesting non-random looking images, it does not express a preference for regularity, coherence, or semantic content. It encourages diverse, but predictive states, so the agent might change many pixels randomly producing images visually similar to noise.  We can use MI between patches of the image $I(top-left, top-right) = H(top-left) - H(top-left|top-right)$ as reward in this case. 

This distinguishes structured images from two trivial solutions:\\ 
\textbf{Blank images}: both patches are predictable, but there is no variation so H(top-left) = 0.\\
\textbf{Independent random noise}: the patches vary, but one does not predict the other. So $H(top-left|top-right) \approx H(top-left)$ again giving 0 mutual information.
\textbf{Structured and variable images}: patches vary across images but share regularities, giving positive mutual information.

The objective can be extended to many random partitions and multiple spatial scales:

\[
R_{\mathrm{structure}}(x)
=
\mathbb E_{(U,V)\sim\mathcal M}
\left[
I(X_U;X_V)
\right],
\]

For discrete pixels or image tokens, one can train:

- a marginal model \(p_\phi(X_V)\), and
- a conditional model \(q_\psi(X_V\mid X_U)\).

A sample-level structure reward is then

\[
r_{\mathrm{structure}}(x)
=
\log q_\psi(x_V\mid x_U)
-
\log p_\phi(x_V).
\]

The first term rewards predictability from context, while the second prevents the predictor from receiving high reward merely because the patch is constant everywhere.

$\log p_\phi(x_V)$ term corresponds to minimum description length(see the section below).

Predictive information can still collapse to a small family of highly regular images—for example, the same checkerboard in every episode. We therefore need to distinguish within-image structure from across-image diversity.

A diversity objective can reward entropy in a global image representation $H(f_\theta(X))$ where f should preferably be a frozen or slowly changing perceptual encoder. Another option is to sample a latent intention $Z$ at the beginning of an episode and maximize $I(C;X_T)$. It is closely related to unsupervised skill discovery: each latent code corresponds to a different controllable mode of image generation.

The two objectives are complementary: predictive information discourages noise generation, global diversity discourages generation of single or a few patterns.

Therefore we could combine these terms into one reward:

$R=αR_empowerment + βR_structure + γR_diversity −λR_cost$

New term $R_cost$ here represents action or complexity cost. This term is needed to avoid useless image modification e.g. changing one pixel black -> white -> black.

This objective does not formally guarantee aesthetically or semantically interesting images. For example, repeated textures, barcodes, or hidden high-frequency signals may score highly despite looking uninteresting to humans.

The term used for a processes that could achieve more and more diverse and complex artifacts is "open-endedness". Our reward combination does not guarantee an open-ended process. Fixed intrinsic objectives can still be exhausted or exploited. Once the agent discovers a finite family of highly controllable, structured images, it may cycle among them indefinitely without producing genuinely new organization. Open-endedness additionally requires a continually expanding space of challenges, or niches — for example through co-evolving competing agents, procedurally generated environments, or learned objectives that change as previous behaviours become common. Possible formalisation is given in \cite{adams2017formal}. 

\subsection{VAE and MDL}
VAE could be used as an approximator for description length. According to Shannon's source coding theorem optimal code length(for a prefix code) that one can assign to a datapoint $x$ is its negative log-likelihood $ - log p(x)$.

VAE models empirical distribution $p_{true}(x)$ as $p_\theta(x) = \int_z p_\theta(x|z)p(z) dz$, approximating the intractable posterior $p_\theta(z|x)$ with the variational posterior $q(z|x)$. For detailed derivation see \cite{kingma2014autoencoding,doersch2016tutorial}.

ELBO objective

\begin{equation}
\begin{aligned}
\mathcal{L}_{\text{ELBO}}(x) &= \log p_\theta(x) - \mathbb{D}_{KL}(q_\phi(z|x)||p_\theta(z|x)) \\
&= \mathbb{E}_{q_{\phi }(z|{}x)}[\log p_{\theta }(x|{}z)]-\mathbb{D}_{KL}(q_{\phi }(z|{}x)\parallel p(z))
\end{aligned}
\end{equation}

Thus we have 

$\log p_\theta(x) ≥ \mathcal{L}_{\text{ELBO}}(x)$

for unknown true distribution this inequality holds in expectation:

$E_{x \sim p_{true}} \log p_\theta(x) ≥ E_{x \sim p_{true}} \mathcal{L}_{\text{ELBO}}(x)$

\section{Problems with Empowerment and DIAYN}\label{problems-with-empowerment-and-diayn}

Single-step empowerment is short-sighted. It can even be zero for obviously easy environments as shown in the xor environment example. Multistep empowerment is better, but it can't be applied without modifications to high-frequency long-term environments like RTS games. On the other hand DIAYN objective is global in a sense that we can sample individual states from a trajectory, and it will provide valid lower bound estimation $I(Z ; single\ state) ≤ I(Z ; whole\ trajectory)$. But empowerment can be zero given one-step or two step estimation.

\begin{figure}[htbp]
\centering
\includegraphics[width=0.48\textwidth]{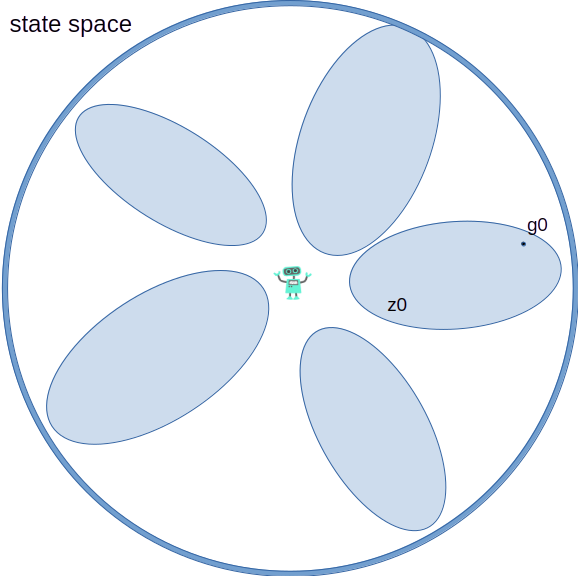}\hfill
\includegraphics[width=0.48\textwidth]{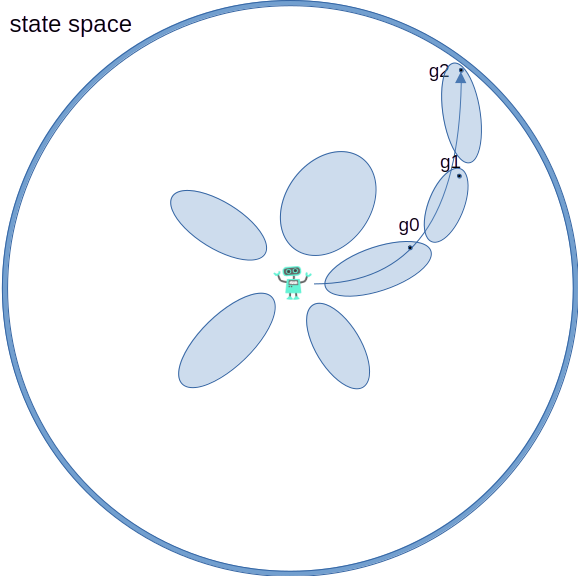}
\caption{State-space partition and traversal. Left: DIAYN partitions the state space into skill-conditioned regions but does not guarantee that an arbitrary goal $g_0$ can be reached; increasing the number of skills may only make these regions thinner. Right: a sequence of reachable goal-conditioned regions $g_0$, $g_1$ and $g_2$ can support traversal through the state space.}
\label{state-space-traversal}
\end{figure}

DIAYN's objective is limited in a sense that it is just for learning skills. Humans or animals use skills to achieve certain goals, skills are combined and used in certain order. DIAYN objective says nothing about how skills should be combined. In more realistic architecture an agent would pursue long-term goals switching between skills as the situation evolves. It's possible to aggregate multiple steps to enable DIAYN to develop different temporal patters, for example achieving the same goal with different locomotion gaits. However it's become very easy for policy to develop pathological solution for skill discrimination e.g. just waiving manipulators in different pattern that are easy to classify. Another example of valid, but uninteresting optimum is in Figure \ref{state-space-traversal}. Assume out agent can move in 2d circle, starting from the centre and observe it's position. It's possible for policy to slice circle along radius and thus recover "skill" vector z while staying near the start. Other rewards such as novelty, e.g. euclidean distance from all states in minibatch could drive agent to explore much larger state-space.
Our example shows that combination of intrinsic rewards can help counter each-others failure modes in some cases.

Issues with empowerment can be addressed by introducing different time scales and state compression. In such case empowerment would be used to reward high-level actions that take many environmental steps to execute.

DIAYN skills can be such low-level actions. And skill selection can be trained by n-step empowerment.

If we choose to use empowerment for only high-level action selection we have to decide which states to use and which to omit. Consider RTS game with 640x480 video frames as observations. Suppose we have high-level actions pursue, build expand, build unit, attack, flee etc. We can't use just pixel observations at skill boundary. Both policy and empowerment estimator would require a feature vector that encodes strategic situations in the game + more detailed description of the current situation, that is exact unit position, recent events such as ``hero just died''.

Multi-layer SFA is a possible front-end to obtain the macro strategic state on which we measure empowerment or DIAYN MI, but exact architecture remains an open research problem.

\section{Research agenda}\label{experiments-options-to-try}

We conclude with discussion of possible experiments intended to test if existing intrinsic motivation methods can produce biological adaptability on different levels.

The proposed direction is related to Schmidhuber's formal theory of creativity and intrinsic motivation, in which agents actively generate experiments and receive intrinsic reward for discovering novel but learnable regularities that improve prediction or data compression \cite{schmidhuber2010formal}.

\subsection{Environment for intrinsic reward evaluation}
Based on previous discussion of MI-based rewards and their "camping on a toy" limitations we propose an environment with natural constraints on camping(or other similarly useless) behaviour. There are two properties of biological environments that counter failure modes of information-based rewards:

\begin{enumerate}
  \item World is adversarial.
  \item Energy is scarce.
\end{enumerate}

We design an environment where agents have an energy level. Analogously to physiological deficits, a low energy level reduces action amplitude and/or makes actions less predictable.

Thus energy level affects the capacity of action to future state channel.

We propose a predator--prey environment with the following observed state:\\
$o_t = (sensor_t, e_t)$ 
where $e_t$ is the energy level.

Agents have a finite storage capacity $e_{max}$ and each action consumes energy:

$e_{t+1} = clip(e_t - c(a_t) + food_t, 0, e_{max})$

Energy level affects actions:

$a_{effective} = m(e_t) a_t + \sigma(e_t) \epsilon_t$

Here $m$ is a magnitude function, might be as simple as $m(e_t)=\max(e_t,m_{\min})$.

The function $\sigma$  determines the magnitude of motor noise and increases as $e_t$ decreases. for example we can define\\
$\sigma(e_t) = \min(\sigma_{max}, - \ln(e_t / e_{max}))$ \\
$\epsilon_t \sim \mathcal{N}(0, I)$

Finding food or capturing prey increases the energy level without providing direct reward signal. 

Note: as discussed earlier continuous actions requires non-zero observation or action uncertainty to keep mutual information finite.

This environment would allow to test directly whether different combinations of intrinsic rewards lead to behaviour that maintains a stable energy level, analogous to energy homeostasis in biological organisms.

Possible tests include environments with:
\begin{enumerate}
	\item controllable toy;
	\item "noisy-tv";
	\item communication channel between predators;
	\item adversarial prey;
\end{enumerate}

Important ablations include disabling the effects of energy on actions and comparing stationary food with adversarial prey. This comparison is necessary because an  intrinsic reward may cause the agent to follow prey for reasons unrelated to energy regulation.\\
Primary metrics to monitor:

\begin{enumerate}
	\item mean energy level
	\item time spent on toys
	\item spatial coverage
	\item energy level at which the agent starts sustained movement towards prey
    \item energy recovery time after reaching the minimum energy level
\end{enumerate}

Interesting extension is test conditions under which communication might develop e.g. when it's hard for an agent to catch the prey on it's own.

\subsection{Recurrent model}

We propose to study a network of recurrent modules where every module is treated as a local agent. Each agent has its own hidden state, observations, incoming messages, actions and intrinsic reward. There is no reward defined for the network as a whole. The modules can communicate during the forward pass, but messages are detached before being passed between agents. Consequently, gradients from one agent cannot propagate through the internal computations of another agent.

For a system of $N$ agents communicating with messages $m$ the shared recurrent update can be written as

\begin{equation}
h_{i,t+1} = F_{\theta}(h_{i,t}, o_{i,t}, m_{i,t}),
\qquad i \in \{1,\ldots,N\},
\end{equation}

where all agents use the same parameters $\theta$, while $h_{i,t}$, $o_{i,t}$ and the position of an agent in the communication graph are different. The shared parameters are analogous to a common genome, while different hidden states, inputs and network positions provide different local contexts. Functional specialisation may therefore emerge without assigning a permanent identity or a separate set of parameters to every module.

Every agent computes an intrinsic reward $r_{i,t}$ only from its own interaction history.

The primary experiment is to test whether the recurrent modules develop stable and complementary roles and whether their joint dynamics exhibit adaptive behaviour that is not explicitly rewarded at the system level.

The basic ablation removes weight sharing. In this condition each module has an independent recurrent function

\begin{equation}
h_{i,t+1} = F_{\theta_i}(h_{i,t}, o_{i,t}, m_{i,t}).
\end{equation}

Independent parameters may make specialisation easier, because different roles can be stored directly in $\theta_i$. Shared weights provide a stronger test: different roles must emerge from local state, experience and network context. As discussed in section \ref{empowerment-in-communication}, a global communication channel might facilitate the formation of collective behaviour. The role of communication can be tested through the following ablations:

\begin{enumerate}
	\item no communication;
	\item local communication;
	\item global low-capacity broadcast;
	\item local + global communication;
\end{enumerate}

This type of agent may be evaluated in a collectively embodied task e.g., sensor + motor policies jointly controlling motion in a maze. In such an environment communication is mandatory. The architecture can also be tested with independently embodied agents. A good example is a predator-prey environment where each recurrent module controls one predator. This type of environment is especially suitable for a communication ablations because the agents can remain independently function when message passing is disabled.

This experiment is not intended to prescribe a final architecture. Its purpose is to test whether local intrinsic objectives, recurrent memory, communication and a shared learning rule are sufficient ingredients for the emergence of functional differentiation and higher-level adaptive organisation.

\subsection{Automatic curriculum in a two-level world model agent}

The limitations discussed in Section~\ref{problems-with-empowerment-and-diayn} suggest an experiment in which motor control and long-term goal selection are learned by separate agents potentially operating at different time scales. The proposed architecture consists of a low-level agent, a high-level agent and a world model. The world model supplies learned state representations and intrinsic signals for training both agents.

Training starts with the low-level agent acting without a valid goal. Its reward is a mixture of learning progress, prediction error and diversity in the learned embedding space. This stage has two purposes: to collect diverse experience for the world model and to train a low-level policy capable of producing non-trivial transitions before it is asked to follow goals. For initial goal conditioning training obvious choices are hindsight training and virtual hindsight training on wm-generated episodes once wm is stable.

After this initial stage, the high-level agent generates a target embedding $g_t$. The low-level policy receives both the current representation $s_t$ and the target $g_t$, and is rewarded for reducing their distance. A simple progress reward is

\begin{equation}
r^{\mathrm{low}}_t = d(z_t,g_t)-d(z_{t+1},g_t).
\end{equation}

The high-level action is held for several environment steps(determined by separate switch head), so selecting one target embedding corresponds to a temporally extended action. The two agents can therefore learn different functions: the low-level agent learns how to realise changes in the representation space, while the high-level agent learns which changes are informative, reachable and useful for continued exploration.

The target space should discard high-frequency details that cannot be controlled over the selected horizon. Slowly varying representations, including representations obtained with SFA-like objectives, are a possible goal space. Long-term empowerment may then be estimated between high-level actions and future states in this compressed representation rather than between individual motor commands and raw observations.

Overall process should achieve state-space traversal similar to one in Figure \ref{state-space-traversal}.

This training process resembles developmental motor learning: initially unstructured self-generated actions establish sensorimotor regularities, after which achieved outcomes can become goals for increasingly directed behaviour. Motor competence and goal selection may then form a coupled curriculum in which each expands the  learning opportunities of the other.

This experiment tests whether temporal hierarchy addresses two complementary limitations. Skill-discovery objectives such as DIAYN do not specify how independently learned skills should be ordered, while short-horizon empowerment does not represent consequences separated from motor actions by many environment steps. A two-level agent instead treats goal-conditioned behaviour as the low-level action space of a slower decision process.

\paragraph{Code availability}
Code for the proposed experiments will be published as the implementations are developed at \url{https://github.com/noskill/reinf}.

\section{Mathematical Appendix}
\label{math-appendix}

\subsubsection{Chain rule for Mutual information}
\begin{equation}
	I(X, Y ; Z) = I(X ; Z) + I(Y ; Z \mid X)
	\label{chain-rule-for-mutual-information-ix-y-z-ix-z-iy-z-x}
\end{equation}

By definition we have  

$I(X, Y ; Z) = H(Z) - H(Z \mid X, Y)$

Expand $I(X ; Z)$ and $I(Y ; Z \mid X)$ to entropies:

$I(X ; Z) = H(Z) - H(Z \mid X)$

$I(Y ; Z \mid X) = H(Z \mid X) - H(Z \mid X, Y)$

Compute sum:

$I(X ; Z) + I(Y ; Z \mid X) = H(Z) - H(Z \mid X) + H(Z \mid X) - H(Z \mid X, Y)$

$I(X ; Z) + I(Y ; Z \mid X) = H(Z) - H(Z \mid X, Y) = I(X, Y ; Z)$

\subsubsection{Donsker-Varadhan MI}\label{donsker-varadhan-mi}

We are starting from KL divergence $D_{KL}(P || Q) =  \int_{}^{}p(u) log \frac{p(u)}{q(u)}du$

\textbf{Step 1. Define Gibbs density.}

Let q(u) be a probability density and T(u) be our arbitrary function. We create a new valid probability density, $g(u)$, by weighting q(u) with $e^{T(u)}$. To ensure g(u) integrates to 1, we must divide by a normalizing constant Z:

\begin{quote}
	$g(u)\  =  \frac{e^{T(u)}q(u)\ }{Z}$
\end{quote}

Where the constant Z is just the expected value over the distribution Q:

$Z = \int_{}^{}e^{T(u)}q(u) du =  E_{Q} e^{T(u)}$

\textbf{Step 2: Use the Non-Negativity of KL Divergence}

$D_{KL}(P || G)  =  \int_{}^{}p(u) log \frac{p(u)}{g(u)}dx \geq 0$

\textbf{Step 3: Multiply by} $\frac{q(u)\ }{q(u)}$

$log \frac{p(u)}{g(u)} =  log \frac{p(u)}{q(u)}  \frac{q(u)}{g(u)} =  log \frac{p(u)}{q(u)} + log \frac{q(u)}{g(u)}$

$g(u)\  = \ \frac{e^{T(u)}q(u)\ }{Z}\  = > \ \frac{g(u)\ }{q(u)}\  = \frac{e^{T(u)}\ }{Z}\ $

$\frac{q(u)\ }{g(u)}\  = \frac{Z}{e^{T(u)}}$

$log\ \frac{q(u)\ }{g(u)}\  = log\ \frac{Z}{e^{T(u)}}\  = \ log\ Z\  - T(u)\ $

\textbf{Step 4: Substitute and Solve}

$\int_{}^{}p(u)\ log\ \frac{p(u)}{q(u)}\frac{q(u) }{g(u)}du  = \int_{}^{}p(u) \big( \log \frac{p(u)}{q(u)}  + \log Z - T(u) \big) du  \geq 0 $

First term $\int_{}^{} p(u) \log \frac{p(u)}{q(u)}du$ is $D_{KL}(P\ ||\ Q)$

Second term $\int_{}^{}p(u) \log Z du  = \log Z \int_{}^{}p(u) du =  \log Z * 1  =  \log Z$

Third term $ \int_{}^{}p(u)T(u)\ du\ $ is $E_{P}T(u)$

$\ D_{KL}(P\ ||\ Q)\ \  + \ {log\ E}_{Q}\ e^{T(u)}\  - \ E_{P}\ T(u)\  \geq 0$

$D_{KL}(P\ ||\ Q) \geq E_{P}\ T(u)\  - log\ E_{Q}\ e^{T(u)}\ \ $

By definition:
$I(X;Y)=D_{KL}\big(P(X,Y)\parallel P(X)P(Y)\big)$

We get mutual information(from definition) with P having density p(x, y) - joint density; and Q having density p(x)p(y) - the product of the marginal densities:

$I(X;\ Y)\  \geq \ E_{P(X,Y)}T(x,y)\  - \ log\ E_{P(X)P(Y)}\ e^{T(x,y)}$

We got lower bound for KL and mutual information.

We can optimise this estimator by finding better function $T$ e.g. with gradient ascent.

\phantomsection
\addcontentsline{toc}{section}{References}
\bibliographystyle{elsarticle-num-names}
\bibliography{references}
\end{document}